%% file: main.tex
\documentclass{article}

\usepackage{microtype}
\usepackage{graphicx}
\usepackage{subcaption}
\usepackage{booktabs}
\usepackage{multirow}
\usepackage{xcolor}
\usepackage{colortbl}
\usepackage{enumitem}
\definecolor{bestbg}{HTML}{FFD1D1}    
\definecolor{secondbg}{HTML}{D1E8FF}  
\definecolor{rowhighlight}{gray}{0.96} 

\usepackage{hyperref}

\usepackage[preprint]{icml2026}

\usepackage{amsmath}
\usepackage{amssymb}
\usepackage{mathtools}
\usepackage{amsthm}

\usepackage[capitalize,noabbrev]{cleveref}

\theoremstyle{plain}
\newtheorem{theorem}{Theorem}[section]

\theoremstyle{definition}

\newtheorem{assumption}[theorem]{Assumption}
\theoremstyle{remark}

\usepackage{booktabs}
\usepackage{multirow}
\usepackage{adjustbox}

\usepackage{makecell}
\usepackage[table]{xcolor}
\usepackage{subcaption}

\usepackage{array}
\newcommand{\rk}[1]{^{\scriptscriptstyle(#1)}}

\usepackage{newfloat}
\usepackage{listings}
\DeclareCaptionStyle{ruled}{labelfont=normalfont,labelsep=colon,strut=off} 
\floatstyle{ruled}
\newfloat{listing}{tb}{lst}{}
\floatname{listing}{Listing}

\usepackage{booktabs}

\usepackage[textsize=tiny]{todonotes}

\icmltitlerunning{Beyond MSE: Rethinking the Evaluation Metric 
and Benchmarking for Irregular Time Series Forecasting}

\begin{document}

\twocolumn[
  \icmltitle{Beyond MSE: Rethinking the Evaluation Metric \\
and Benchmarking for Irregular Time Series Forecasting}



  \icmlsetsymbol{equal}{*}

  \begin{icmlauthorlist}
    \icmlauthor{Rongwen Li}{hnu}
    \icmlauthor{Haixin Xie}{hnu}
    \icmlauthor{Xiao Wang}{hnu}
    \icmlauthor{Changjian Chen}{hnu}
  \end{icmlauthorlist}

  \icmlaffiliation{hnu}{College of Computer Science and Electronic Engineering, Hunan University, Changsha, Hunan, China}

  \icmlcorrespondingauthor{Changjian Chen}{changjianchen@hnu.edu.cn}

  \icmlkeywords{Machine Learning, ICML}

  \vskip 0.3in
]



\printAffiliationsAndNotice{}  

\begin{abstract}
Existing research on irregular time-series forecasting has primarily focused on model design, while evaluation metrics remain insufficiently studied. Existing benchmarks typically use mean squared error (MSE) as the evaluation metric. We show that, in irregular forecasting, MSE is determined not only by the model prediction but also by the sample-specific timestamp sampling distributions, leading to a biased assessment of the models’ continuous-time predictive performance. To address this issue, we propose the \textit{Continuous-time Squared Error (CSE)}, which employs importance weighting to eliminate the influence of the timestamp sampling distributions. We further theoretically prove that CSE's asymptotic estimation error with respect to continuous-time risk is no greater than that of MSE. Finally, we construct a systematic benchmark covering synthetic, semi-synthetic, and eight real-world datasets to validate the effectiveness of CSE and systematically evaluate models’ continuous-time predictive performance. Experiments show that CSE can recover continuous-time risk more accurately than MSE, while relying solely on MSE may not fully reflect models’ continuous-time predictive performance in real-world scenarios. Our code can be obtained at https://github.com/hnu-vis/ITS-Bench.
\end{abstract}

\input{paper/introduction}

\input{paper/related_work}

\input{paper/preliminary}

\input{paper/method}

\input{paper/experiments}

\input{paper/conclusion}


\nocite{langley00}

\bibliography{main}
\bibliographystyle{icml2026}

\newpage
\appendix

\section{Proof of Theorem~1}
\label{app:proof}

For the $s$-th trajectory, denote
\begin{equation}
p_s(t)=p(t\mid X_s),
\qquad
\ell_s(t)=\ell(t,X_s),
\end{equation}
and define
\begin{align}
r_{\mathrm{ct}}(X_s)
&=
\frac{1}{|\mathcal T|}
\int_{\mathcal T}
\ell_s(t)\,dt,\\
r_{\mathrm{obs}}(X_s)
&=
\int_{\mathcal T}
\ell_s(t)p_s(t)\,dt.
\end{align}
Accordingly,
\begin{equation}
R_{\mathrm{ct}}
=
\mathbb E_X[r_{\mathrm{ct}}(X)],
\qquad
R_{\mathrm{obs}}
=
\mathbb E_X[r_{\mathrm{obs}}(X)].
\end{equation}

We first introduce the oracle importance-sampling terms
\begin{equation}
A_s
=
\frac{1}{L_s}
\sum_{i=1}^{L_s}
\frac{\ell_s(t_{s,i})}{p_s(t_{s,i})},
\qquad
B_s
=
\frac{1}{L_s}
\sum_{i=1}^{L_s}
\frac{1}{p_s(t_{s,i})},
\end{equation}
and their estimated counterparts
\begin{equation}
\widehat A_s
=
\frac{1}{L_s}
\sum_{i=1}^{L_s}
\frac{\ell_s(t_{s,i})}
{\widehat p(t_{s,i}\mid X_s)},
\qquad
\widehat B_s
=
\frac{1}{L_s}
\sum_{i=1}^{L_s}
\frac{1}
{\widehat p(t_{s,i}\mid X_s)}.
\end{equation}

By Assumption~1,
$p_s(t)\geq\rho>0$ and
$\widehat p(t\mid X_s)$ is uniformly consistent on
$\mathcal T$. Hence, almost surely, for sufficiently large
$L_s$,
\begin{equation}
\inf_{t\in\mathcal T}
\widehat p(t\mid X_s)
\geq \frac{\rho}{2},
\end{equation}
and therefore
\begin{equation}
\begin{aligned}
\sup_{t\in\mathcal T}
\left|
\frac{1}{\widehat p(t\mid X_s)}
-
\frac{1}{p_s(t)}
\right|
&\leq
\frac{2}{\rho^2}
\sup_{t\in\mathcal T}
\left|
\widehat p(t\mid X_s)-p_s(t)
\right|\\
&\xrightarrow{\mathrm{a.s.}}0.
\end{aligned}
\label{eq:inverse_density_consistency}
\end{equation}

Let $M<\infty$ satisfy
$0\leq\ell(t,X)\leq M$. It follows that
\begin{align}
\left|\widehat A_s-A_s\right|
&\leq
M
\sup_{t\in\mathcal T}
\left|
\frac{1}{\widehat p(t\mid X_s)}
-
\frac{1}{p_s(t)}
\right|
\xrightarrow{\mathrm{a.s.}}0,\\
\left|\widehat B_s-B_s\right|
&\leq
\sup_{t\in\mathcal T}
\left|
\frac{1}{\widehat p(t\mid X_s)}
-
\frac{1}{p_s(t)}
\right|
\xrightarrow{\mathrm{a.s.}}0.
\end{align}

Conditional on $X_s$, the future timestamps are independently
and identically distributed according to $p_s(t)$. The strong
law of large numbers therefore gives
\begin{align}
A_s
&\xrightarrow{\mathrm{a.s.}}
\mathbb E_{T\mid X_s}
\left[
\frac{\ell_s(T)}{p_s(T)}
\right]
=
\int_{\mathcal T}\ell_s(t)\,dt,\\
B_s
&\xrightarrow{\mathrm{a.s.}}
\mathbb E_{T\mid X_s}
\left[
\frac{1}{p_s(T)}
\right]
=
|\mathcal T|.
\end{align}
Combining the above results yields
\begin{equation}
\widehat A_s
\xrightarrow{\mathrm{a.s.}}
\int_{\mathcal T}\ell_s(t)\,dt,
\qquad
\widehat B_s
\xrightarrow{\mathrm{a.s.}}
|\mathcal T|.
\end{equation}

Define the trajectory-level CSE estimator as
\begin{equation}
\widehat r_{\mathrm{ct},s}
=
\frac{
\sum_{i=1}^{L_s}
\ell_s(t_{s,i})/
\widehat p(t_{s,i}\mid X_s)
}{
\sum_{i=1}^{L_s}
1/\widehat p(t_{s,i}\mid X_s)
}
=
\frac{\widehat A_s}{\widehat B_s}.
\end{equation}
Since $|\mathcal T|>0$, the continuous mapping theorem gives
\begin{equation}
\widehat r_{\mathrm{ct},s}
\xrightarrow{\mathrm{a.s.}}
\frac{1}{|\mathcal T|}
\int_{\mathcal T}\ell_s(t)\,dt
=
r_{\mathrm{ct}}(X_s).
\label{eq:trajectory_cse_limit}
\end{equation}

For MSE, the conditional strong law directly gives
\begin{equation}
\widehat r_{\mathrm{obs},s}
=
\frac{1}{L_s}
\sum_{i=1}^{L_s}
\ell_s(t_{s,i})
\xrightarrow{\mathrm{a.s.}}
\mathbb E_{T\mid X_s}[\ell_s(T)]
=
r_{\mathrm{obs}}(X_s).
\label{eq:trajectory_mse_limit}
\end{equation}

Let
$\underline L=\min_{1\leq s\leq S}L_s$.
For the joint limit below, Assumption~1 is understood to hold
uniformly over the test trajectories. Consequently,
Eqs.~\eqref{eq:trajectory_cse_limit} and
\eqref{eq:trajectory_mse_limit} imply
\begin{align}
\frac{1}{S}
\sum_{s=1}^{S}
\left|
\widehat r_{\mathrm{ct},s}
-
r_{\mathrm{ct}}(X_s)
\right|
&\xrightarrow{\mathrm{a.s.}}0,\\
\frac{1}{S}
\sum_{s=1}^{S}
\left|
\widehat r_{\mathrm{obs},s}
-
r_{\mathrm{obs}}(X_s)
\right|
&\xrightarrow{\mathrm{a.s.}}0
\end{align}
as $\underline L\rightarrow\infty$.

Since the test trajectories are mutually independent and the
trajectory-level risks are bounded, another application of the
strong law yields
\begin{equation}
\frac{1}{S}
\sum_{s=1}^{S}
r_{\mathrm{ct}}(X_s)
\xrightarrow{\mathrm{a.s.}}
R_{\mathrm{ct}},
\qquad
\frac{1}{S}
\sum_{s=1}^{S}
r_{\mathrm{obs}}(X_s)
\xrightarrow{\mathrm{a.s.}}
R_{\mathrm{obs}}.
\end{equation}
\begin{table*}[t]
\centering
\scriptsize
\setlength{\tabcolsep}{1.8pt}
\renewcommand{\arraystretch}{1.15}
\caption{Complete MSE and CSE results on CESNET, FNSPID, GDELT, and HumanActivity. Values are mean $\pm$ standard deviation over three random seeds; standard deviations are rounded to three decimal places.}
\label{tab:app_mse_cse_part1}
\begin{adjustbox}{max width=\textwidth}
\begin{tabular}{l*{4}{cc}}
\toprule
Model & \multicolumn{2}{c}{\shortstack{CESNET\\$\times10^{-1}$}} & \multicolumn{2}{c}{\shortstack{FNSPID\\$\times10^{-1}$}} & \multicolumn{2}{c}{\shortstack{GDELT\\$\times10^{0}$}} & \multicolumn{2}{c}{\shortstack{HumanActivity\\$\times10^{-2}$}} \\
\cmidrule(lr){2-3}\cmidrule(lr){4-5}\cmidrule(lr){6-7}\cmidrule(lr){8-9}
& MSE & CSE & MSE & CSE & MSE & CSE & MSE & CSE \\
\midrule
APN & $8.76\!\pm\!0.007^{(5)}$ & $8.58\!\pm\!0.008^{(5)}$ & $\underline{2.38}\!\pm\!0.143^{(2)}$ & $\underline{2.38}\!\pm\!0.144^{(2)}$ & $\mathbf{1.02}\!\pm\!0.001^{(1)}$ & $\mathbf{0.99}\!\pm\!0.001^{(1)}$ & $5.68\!\pm\!0.035^{(6)}$ & $5.65\!\pm\!0.039^{(6)}$ \\
ASTGI & $\mathbf{8.47}\!\pm\!0.137^{(1)}$ & $\mathbf{8.26}\!\pm\!0.122^{(1)}$ & $3.28\!\pm\!0.213^{(6)}$ & $3.25\!\pm\!0.211^{(6)}$ & $1.10\!\pm\!0.003^{(4)}$ & $1.04\!\pm\!0.003^{(4)}$ & $5.46\!\pm\!0.167^{(3)}$ & $5.41\!\pm\!0.167^{(3)}$ \\
GRU-D & $10.22\!\pm\!0.464^{(10)}$ & $9.91\!\pm\!0.466^{(10)}$ & $5.24\!\pm\!0.800^{(8)}$ & $5.20\!\pm\!0.791^{(8)}$ & $1.22\!\pm\!0.003^{(10)}$ & $1.15\!\pm\!0.003^{(10)}$ & $8.56\!\pm\!1.497^{(9)}$ & $8.56\!\pm\!1.515^{(9)}$ \\
GraFITi & $\underline{8.50}\!\pm\!0.113^{(2)}$ & $\underline{8.28}\!\pm\!0.102^{(2)}$ & $3.21\!\pm\!0.755^{(4)}$ & $3.22\!\pm\!0.760^{(4)}$ & $1.09\!\pm\!0.002^{(3)}$ & $\underline{1.01}\!\pm\!0.002^{(2)}$ & $\underline{5.44}\!\pm\!0.038^{(2)}$ & $\underline{5.40}\!\pm\!0.034^{(2)}$ \\
HyperIMTS & $8.90\!\pm\!0.073^{(7)}$ & $8.65\!\pm\!0.060^{(7)}$ & $2.72\!\pm\!0.593^{(3)}$ & $2.71\!\pm\!0.592^{(3)}$ & $1.18\!\pm\!0.002^{(8)}$ & $1.09\!\pm\!0.002^{(8)}$ & $\mathbf{5.42}\!\pm\!0.037^{(1)}$ & $\mathbf{5.38}\!\pm\!0.039^{(1)}$ \\
KAFNet & $8.73\!\pm\!0.012^{(4)}$ & $8.54\!\pm\!0.003^{(4)}$ & $\mathbf{2.23}\!\pm\!0.047^{(1)}$ & $\mathbf{2.22}\!\pm\!0.043^{(1)}$ & $\underline{1.08}\!\pm\!0.005^{(2)}$ & $1.02\!\pm\!0.005^{(3)}$ & $5.62\!\pm\!0.021^{(4)}$ & $5.58\!\pm\!0.025^{(4)}$ \\
NeuralFlow & $9.98\!\pm\!0.103^{(8)}$ & $9.69\!\pm\!0.118^{(8)}$ & $7.33\!\pm\!0.138^{(10)}$ & $7.29\!\pm\!0.140^{(10)}$ & $1.18\!\pm\!0.005^{(9)}$ & $1.09\!\pm\!0.005^{(9)}$ & $17.53\!\pm\!2.512^{(11)}$ & $17.51\!\pm\!2.523^{(11)}$ \\
SeFT & $10.11\!\pm\!0.183^{(9)}$ & $9.79\!\pm\!0.177^{(9)}$ & $8.70\!\pm\!0.958^{(11)}$ & $8.66\!\pm\!0.957^{(11)}$ & $1.13\!\pm\!0.003^{(6)}$ & $1.05\!\pm\!0.003^{(5)}$ & $9.38\!\pm\!0.422^{(10)}$ & $9.36\!\pm\!0.406^{(10)}$ \\
Warpformer & $8.58\!\pm\!0.067^{(3)}$ & $8.37\!\pm\!0.058^{(3)}$ & $3.66\!\pm\!0.216^{(7)}$ & $3.63\!\pm\!0.220^{(7)}$ & $1.30\!\pm\!0.016^{(11)}$ & $1.20\!\pm\!0.016^{(11)}$ & $5.65\!\pm\!0.143^{(5)}$ & $5.61\!\pm\!0.141^{(5)}$ \\
mTAN & $10.47\!\pm\!0.153^{(11)}$ & $10.14\!\pm\!0.106^{(11)}$ & $6.13\!\pm\!0.743^{(9)}$ & $6.09\!\pm\!0.738^{(9)}$ & $1.12\!\pm\!0.003^{(5)}$ & $1.05\!\pm\!0.003^{(6)}$ & $7.33\!\pm\!0.108^{(8)}$ & $7.30\!\pm\!0.107^{(8)}$ \\
tPatchGNN & $8.86\!\pm\!0.140^{(6)}$ & $8.61\!\pm\!0.114^{(6)}$ & $3.26\!\pm\!0.269^{(5)}$ & $3.24\!\pm\!0.269^{(5)}$ & $1.15\!\pm\!0.006^{(7)}$ & $1.08\!\pm\!0.006^{(7)}$ & $5.80\!\pm\!0.121^{(7)}$ & $5.76\!\pm\!0.132^{(7)}$ \\
\bottomrule
\end{tabular}
\end{adjustbox}
\end{table*}

Therefore, as $S,\underline L\rightarrow\infty$,
\begin{equation}
\mathrm{CSE}
\xrightarrow{\mathrm{a.s.}}
R_{\mathrm{ct}},
\qquad
\mathrm{MSE}
\xrightarrow{\mathrm{a.s.}}
R_{\mathrm{obs}}.
\end{equation}
It follows that
\begin{align}
\lim_{S,\underline L\rightarrow\infty}
\left|
\mathrm{CSE}-R_{\mathrm{ct}}
\right|
&=0,\\
\lim_{S,\underline L\rightarrow\infty}
\left|
\mathrm{MSE}-R_{\mathrm{ct}}
\right|
&=
\left|
R_{\mathrm{obs}}-R_{\mathrm{ct}}
\right|
\end{align}
almost surely. Hence,
\begin{equation}
\lim_{S,\underline L\rightarrow\infty}
\left|
\mathrm{CSE}-R_{\mathrm{ct}}
\right|
\leq
\lim_{S,\underline L\rightarrow\infty}
\left|
\mathrm{MSE}-R_{\mathrm{ct}}
\right|.
\end{equation}

When
$p(t\mid X)=1/|\mathcal T|$ almost everywhere,
we have
$R_{\mathrm{obs}}=R_{\mathrm{ct}}$, and both asymptotic
estimation errors are zero. More generally, equality holds
whenever $R_{\mathrm{obs}}=R_{\mathrm{ct}}$, whereas the
inequality is strict when
$R_{\mathrm{obs}}\neq R_{\mathrm{ct}}$.
The self-normalized estimator may be biased at a finite sample
size; the result above concerns its consistency and asymptotic
estimation error.

\section{Derivation of the Evaluation Discrepancy}
\label{app:decomposition}

Recall that
\begin{equation}
p_T(t)=\mathbb E_X[p(t\mid X)],
\qquad
\overline{\ell}(t)=\mathbb E_X[\ell(t,X)].
\end{equation}

For the sampling-dependence gap,
\begin{align}
G_{\mathrm{samp}}
&=
R_{\mathrm{obs}}-R_{\mathrm{global}}
\nonumber\\
&=
\mathbb E_X
\left[
\int_{\mathcal T}
\ell(t,X)p(t\mid X)\,dt
\right]
-
\mathbb E_X
\left[
\int_{\mathcal T}
\ell(t,X)p_T(t)\,dt
\right]
\nonumber\\
&=
\int_{\mathcal T}
\left\{
\mathbb E_X[\ell(t,X)p(t\mid X)]
-
\mathbb E_X[\ell(t,X)]
\mathbb E_X[p(t\mid X)]
\right\}dt
\nonumber\\
&=
\int_{\mathcal T}
\operatorname{Cov}_X
\left(
\ell(t,X),p(t\mid X)
\right)dt.
\label{eq:app_sampling_gap}
\end{align}

For the temporal-distribution gap, let
$U\sim\operatorname{Unif}(\mathcal T)$. Since
$\mathbb E_U[|\mathcal T|p_T(U)]=1$, we have
\begin{align}
G_{\mathrm{time}}
&=
R_{\mathrm{global}}-R_{\mathrm{ct}}
\nonumber\\
&=
\int_{\mathcal T}
\overline{\ell}(t)p_T(t)\,dt
-
\frac{1}{|\mathcal T|}
\int_{\mathcal T}
\overline{\ell}(t)\,dt
\nonumber\\
&=
\mathbb E_U
\left[
\overline{\ell}(U)|\mathcal T|p_T(U)
\right]
-
\mathbb E_U[\overline{\ell}(U)]
\mathbb E_U[|\mathcal T|p_T(U)]
\nonumber\\
&=
\operatorname{Cov}_{U\sim\operatorname{Unif}(\mathcal T)}
\left(
\overline{\ell}(U),
|\mathcal T|p_T(U)
\right).
\label{eq:app_time_gap}
\end{align}

Therefore,
\begin{equation}
R_{\mathrm{obs}}-R_{\mathrm{ct}}
=
G_{\mathrm{samp}}+G_{\mathrm{time}}.
\end{equation}
Here, $G_{\mathrm{samp}}$ captures the effect of
sample-dependent sampling, whereas $G_{\mathrm{time}}$
captures the effect of population-level temporal
non-uniformity.

\section{Global-Time Squared Error}
\label{app:gse}

The marginal timestamp density is estimated by averaging the
trajectory-specific conditional densities:
\begin{equation}
\widehat p_T(t)
=
\frac{1}{S}
\sum_{r=1}^{S}
\widehat p(t\mid X_r).
\end{equation}

Using $\widehat p(t\mid X_s)$ as the proposal density and
$\widehat p_T(t)$ as the target density, we define the
Global-Time Squared Error as
\begin{equation}
\mathrm{GSE}
=
\frac{1}{S}
\sum_{s=1}^{S}
\frac{
\displaystyle
\sum_{i=1}^{L_s}
\ell(t_{s,i},X_s)
\frac{
\widehat p_T(t_{s,i})
}{
\widehat p(t_{s,i}\mid X_s)
}
}{
\displaystyle
\sum_{i=1}^{L_s}
\frac{
\widehat p_T(t_{s,i})
}{
\widehat p(t_{s,i}\mid X_s)
}
}.
\label{eq:app_gse}
\end{equation}

GSE evaluates each trajectory under the dataset-level marginal
timestamp distribution, thereby removing sample-specific
sampling effects while retaining population-level temporal
non-uniformity. It serves as the empirical counterpart of
$R_{\mathrm{global}}$ and an intermediate reference between
MSE and CSE.

\begin{table*}[t]
\centering
\scriptsize
\setlength{\tabcolsep}{1.8pt}
\renewcommand{\arraystretch}{1.15}
\caption{Complete MSE and CSE results on MIMIC-III, RepoHealth, StudentLife, and USHCN. Values are mean $\pm$ standard deviation over three random seeds; standard deviations are rounded to three decimal places.}
\label{tab:app_mse_cse_part2}
\begin{adjustbox}{max width=\textwidth}
\begin{tabular}{l*{4}{cc}}
\toprule
Model & \multicolumn{2}{c}{\shortstack{MIMIC-III\\$\times10^{-1}$}} & \multicolumn{2}{c}{\shortstack{RepoHealth\\$\times10^{-1}$}} & \multicolumn{2}{c}{\shortstack{StudentLife\\$\times10^{-1}$}} & \multicolumn{2}{c}{\shortstack{USHCN\\$\times10^{-1}$}} \\
\cmidrule(lr){2-3}\cmidrule(lr){4-5}\cmidrule(lr){6-7}\cmidrule(lr){8-9}
& MSE & CSE & MSE & CSE & MSE & CSE & MSE & CSE \\
\midrule
APN & $6.66\!\pm\!0.038^{(6)}$ & $6.72\!\pm\!0.038^{(6)}$ & $\underline{3.36}\!\pm\!0.069^{(2)}$ & $\mathbf{3.57}\!\pm\!0.080^{(1)}$ & $\underline{6.61}\!\pm\!0.009^{(2)}$ & $6.68\!\pm\!0.006^{(3)}$ & $6.04\!\pm\!0.554^{(4)}$ & $6.71\!\pm\!0.686^{(5)}$ \\
ASTGI & $10.50\!\pm\!0.000^{(11)}$ & $10.62\!\pm\!0.000^{(11)}$ & $6.27\!\pm\!2.310^{(7)}$ & $6.53\!\pm\!2.378^{(7)}$ & $6.65\!\pm\!0.036^{(3)}$ & $\underline{6.65}\!\pm\!0.038^{(2)}$ & $5.14\!\pm\!0.169^{(3)}$ & $5.66\!\pm\!0.179^{(3)}$ \\
GRU-D & $6.67\!\pm\!0.043^{(7)}$ & $6.75\!\pm\!0.046^{(7)}$ & $11.68\!\pm\!0.695^{(11)}$ & $11.95\!\pm\!0.672^{(11)}$ & $7.21\!\pm\!0.045^{(8)}$ & $7.23\!\pm\!0.045^{(8)}$ & $7.78\!\pm\!0.690^{(11)}$ & $8.59\!\pm\!0.756^{(11)}$ \\
GraFITi & $\mathbf{5.89}\!\pm\!0.264^{(1)}$ & $\underline{6.02}\!\pm\!0.278^{(2)}$ & $\mathbf{3.32}\!\pm\!0.341^{(1)}$ & $\underline{3.61}\!\pm\!0.342^{(2)}$ & $6.72\!\pm\!0.164^{(6)}$ & $6.70\!\pm\!0.162^{(4)}$ & $\mathbf{4.50}\!\pm\!0.252^{(1)}$ & $\mathbf{4.89}\!\pm\!0.299^{(1)}$ \\
HyperIMTS & $\underline{5.94}\!\pm\!0.068^{(2)}$ & $\mathbf{5.92}\!\pm\!0.068^{(1)}$ & $3.72\!\pm\!0.437^{(3)}$ & $3.97\!\pm\!0.444^{(3)}$ & $\mathbf{6.54}\!\pm\!0.067^{(1)}$ & $\mathbf{6.56}\!\pm\!0.062^{(1)}$ & $6.67\!\pm\!1.837^{(8)}$ & $7.30\!\pm\!2.073^{(8)}$ \\
KAFNet & $6.41\!\pm\!0.021^{(3)}$ & $6.47\!\pm\!0.022^{(3)}$ & $5.72\!\pm\!3.034^{(6)}$ & $5.98\!\pm\!3.061^{(6)}$ & $6.66\!\pm\!0.023^{(4)}$ & $6.72\!\pm\!0.024^{(5)}$ & $7.14\!\pm\!0.418^{(9)}$ & $7.82\!\pm\!0.406^{(9)}$ \\
NeuralFlow & $7.05\!\pm\!0.042^{(8)}$ & $7.19\!\pm\!0.044^{(9)}$ & $10.13\!\pm\!0.283^{(10)}$ & $10.44\!\pm\!0.320^{(10)}$ & $7.83\!\pm\!0.218^{(10)}$ & $7.86\!\pm\!0.212^{(10)}$ & $6.07\!\pm\!0.720^{(5)}$ & $6.74\!\pm\!0.764^{(6)}$ \\
SeFT & $7.10\!\pm\!0.060^{(9)}$ & $7.15\!\pm\!0.060^{(8)}$ & $9.13\!\pm\!0.454^{(8)}$ & $9.44\!\pm\!0.463^{(8)}$ & $8.94\!\pm\!1.706^{(11)}$ & $8.95\!\pm\!1.700^{(11)}$ & $7.64\!\pm\!0.137^{(10)}$ & $8.41\!\pm\!0.170^{(10)}$ \\
Warpformer & $6.42\!\pm\!0.041^{(4)}$ & $6.50\!\pm\!0.045^{(5)}$ & $5.04\!\pm\!0.555^{(5)}$ & $5.33\!\pm\!0.598^{(5)}$ & $6.78\!\pm\!0.065^{(7)}$ & $6.80\!\pm\!0.066^{(7)}$ & $6.13\!\pm\!0.168^{(6)}$ & $6.67\!\pm\!0.195^{(4)}$ \\
mTAN & $7.47\!\pm\!0.034^{(10)}$ & $7.54\!\pm\!0.032^{(10)}$ & $10.09\!\pm\!0.438^{(9)}$ & $10.39\!\pm\!0.443^{(9)}$ & $7.32\!\pm\!0.069^{(9)}$ & $7.34\!\pm\!0.067^{(9)}$ & $\underline{4.84}\!\pm\!0.174^{(2)}$ & $\underline{5.23}\!\pm\!0.232^{(2)}$ \\
tPatchGNN & $6.44\!\pm\!0.029^{(5)}$ & $6.48\!\pm\!0.021^{(4)}$ & $4.55\!\pm\!0.322^{(4)}$ & $4.82\!\pm\!0.335^{(4)}$ & $6.68\!\pm\!0.059^{(5)}$ & $6.73\!\pm\!0.060^{(6)}$ & $6.36\!\pm\!1.306^{(7)}$ & $7.05\!\pm\!1.384^{(7)}$ \\
\bottomrule
\end{tabular}
\end{adjustbox}
\end{table*}

\section{Benchmark Details}
\label{app:benchmark_details}

\subsection{Synthetic and Semi-Synthetic Data}
\label{app:controlled_data}

\paragraph{Common setup.}
The controlled benchmark contains two fully synthetic datasets,
Synthetic-Regime and Synthetic-Multiscale, and two
semi-synthetic datasets constructed from ETTm1 and Weather.
Each dataset contains 512 training, 128 validation, and 128 test
trajectories. For each trajectory, 16 historical observations are
provided as input. Training and validation use 30 future target
points, whereas each test realization contains 128 future query
timestamps. For fully synthetic data, the historical and future
intervals are $[0,0.5]$ and $(0.5,1]$, respectively. Gaussian
noise with standard deviation $0.002$ is added only to the
historical observations.

\paragraph{Synthetic-Regime.}
Each trajectory is controlled by
$z_1,z_2\overset{\mathrm{i.i.d.}}{\sim}\mathcal{N}(0,1)$.
We first define
\begin{equation}
\begin{aligned}
r(t)
&=
\operatorname{clip}
\left(
\frac{t-0.58}{0.68-0.58},0,1
\right),\\
w(t)
&=
r(t)^2\{3-2r(t)\}.
\end{aligned}
\label{eq:app_regime_weight}
\end{equation}
The smooth and periodic components are
\begin{equation}
\begin{aligned}
b_1(t)&=0.10z_1+0.015\sin(2\pi t),\\
b_2(t)&=0.10z_2+0.012\cos(2\pi t),\\
b_3(t)&=0.07(z_1-z_2)+0.010\sin(4\pi t),\\
p_1(t)&=(0.48+0.04z_2)\sin\!\left(12\pi(t-0.64)\right),\\
p_2(t)&=(0.44+0.04z_1)\cos\!\left(10\pi(t-0.64)\right),\\
p_3(t)&=\{0.40+0.03(z_1+z_2)\}
        \sin\!\left(8\pi(t-0.64)\right).
\end{aligned}
\label{eq:app_regime_components}
\end{equation}
The final three-dimensional trajectory is
\begin{equation}
y_j(t)=b_j(t)+w(t)p_j(t),
\qquad j\in\{1,2,3\}.
\label{eq:app_regime_trajectory}
\end{equation}
The transition function $w(t)$ continuously changes the process
from a low-amplitude smooth regime to a high-frequency periodic
regime, producing different levels of forecasting difficulty
across the future interval.

\paragraph{Synthetic-Multiscale.}
For
$z_1,z_2,z_3\overset{\mathrm{i.i.d.}}{\sim}\mathcal{N}(0,1)$,
we define
\begin{equation}
\begin{aligned}
g_0(t)
&=
0.25z_1+0.18t
+0.32\sin(2\pi\!\cdot\!1.25t+0.20z_2),\\
g_h(t)
&=
0.14(1+0.08z_1)
\sin(2\pi\!\cdot\!8t+0.25z_3),\\
\mu
&=
0.72+0.025\tanh(z_3),\\
g_b(t)
&=
0.30(1+0.06z_2)
\exp\!\left\{
-\frac{(t-\mu)^2}{2(0.055)^2}
\right\}.
\end{aligned}
\label{eq:app_multiscale_components}
\end{equation}
The output channels are
\begin{equation}
\begin{aligned}
y_1(t)
&=
g_0(t)+g_h(t)+g_b(t),\\
y_2(t)
&=
0.75g_0(t)-0.80g_h(t)+0.65g_b(t)\\
&\quad+0.08\cos(5\pi t),\\
y_3(t)
&=
-0.45g_0(t)+0.55g_h(t)-0.70g_b(t)\\
&\quad+0.12t
+0.10\sin(1.5\pi t+0.30z_1).
\end{aligned}
\label{eq:app_multiscale_trajectory}
\end{equation}
This construction combines trend, low- and high-frequency
periodicity, and a localized transient component.

\paragraph{Semi-synthetic datasets.}
For ETTm1 and Weather, the original regularly sampled series are
chronologically divided into training, validation, and test
segments with a ratio of $60\%/20\%/20\%$. Standardization
statistics are fitted using the training segment. Each sample
contains 96 historical grid points and 512 future grid points.
We retain 16 historical points as input and sample 30 future
points for training and validation. At test time, 128 future
points are selected from the complete 512-point future grid,
which is retained to compute the reference continuous-time risk.

For all four datasets, test timestamps are sampled from
\begin{equation}
p_{\alpha}(u)
=
\alpha\,\operatorname{Beta}(u;10,60)
+
(1-\alpha)\,\operatorname{Uniform}(0,1),
\label{eq:app_sampling_distribution}
\end{equation}
where
$\alpha\in\{0,0.3,0.5,0.9,0.99\}$ controls the degree of
temporal non-uniformity. For each $\alpha$, the trained model,
historical observations, and underlying future trajectory are
fixed, while only the test query timestamps are resampled.
Each setting is evaluated using 20 timestamp resamples shared
across all models.

\begin{table*}[t]
\centering
\scriptsize
\setlength{\tabcolsep}{1.8pt}
\renewcommand{\arraystretch}{1.15}
\caption{Complete MAE and CAE results on CESNET, FNSPID, GDELT, and HumanActivity. Values are mean $\pm$ standard deviation over three random seeds; standard deviations are rounded to three decimal places.}
\label{tab:app_mae_cae_part1}
\begin{adjustbox}{max width=\textwidth}
\begin{tabular}{l*{4}{cc}}
\toprule
Model & \multicolumn{2}{c}{\shortstack{CESNET\\$\times10^{-1}$}} & \multicolumn{2}{c}{\shortstack{FNSPID\\$\times10^{-1}$}} & \multicolumn{2}{c}{\shortstack{GDELT\\$\times10^{-1}$}} & \multicolumn{2}{c}{\shortstack{HumanActivity\\$\times10^{-1}$}} \\
\cmidrule(lr){2-3}\cmidrule(lr){4-5}\cmidrule(lr){6-7}\cmidrule(lr){8-9}
& MAE & CAE & MAE & CAE & MAE & CAE & MAE & CAE \\
\midrule
APN & $6.69\!\pm\!0.003^{(5)}$ & $6.62\!\pm\!0.003^{(5)}$ & $\underline{1.86}\!\pm\!0.173^{(2)}$ & $\underline{1.85}\!\pm\!0.173^{(2)}$ & $\mathbf{6.59}\!\pm\!0.066^{(1)}$ & $\mathbf{6.50}\!\pm\!0.067^{(1)}$ & $1.43\!\pm\!0.031^{(6)}$ & $1.42\!\pm\!0.032^{(6)}$ \\
ASTGI & $\mathbf{6.55}\!\pm\!0.064^{(1)}$ & $\mathbf{6.48}\!\pm\!0.067^{(1)}$ & $2.19\!\pm\!0.092^{(6)}$ & $2.18\!\pm\!0.093^{(6)}$ & $6.69\!\pm\!0.088^{(4)}$ & $6.60\!\pm\!0.088^{(4)}$ & $1.38\!\pm\!0.024^{(3)}$ & $1.38\!\pm\!0.024^{(3)}$ \\
GRU-D & $7.45\!\pm\!0.249^{(10)}$ & $7.37\!\pm\!0.262^{(10)}$ & $2.76\!\pm\!0.286^{(8)}$ & $2.75\!\pm\!0.284^{(8)}$ & $6.81\!\pm\!0.033^{(10)}$ & $6.72\!\pm\!0.032^{(10)}$ & $2.06\!\pm\!0.251^{(9)}$ & $2.06\!\pm\!0.253^{(9)}$ \\
GraFITi & $\underline{6.56}\!\pm\!0.033^{(2)}$ & $\underline{6.49}\!\pm\!0.033^{(2)}$ & $1.96\!\pm\!0.097^{(4)}$ & $1.95\!\pm\!0.096^{(4)}$ & $6.66\!\pm\!0.146^{(3)}$ & $6.57\!\pm\!0.146^{(3)}$ & $\underline{1.37}\!\pm\!0.012^{(2)}$ & $\underline{1.37}\!\pm\!0.013^{(2)}$ \\
HyperIMTS & $6.79\!\pm\!0.056^{(7)}$ & $6.75\!\pm\!0.063^{(7)}$ & $1.91\!\pm\!0.391^{(3)}$ & $1.91\!\pm\!0.390^{(3)}$ & $6.74\!\pm\!0.072^{(8)}$ & $6.65\!\pm\!0.074^{(8)}$ & $\mathbf{1.36}\!\pm\!0.013^{(1)}$ & $\mathbf{1.36}\!\pm\!0.013^{(1)}$ \\
KAFNet & $6.68\!\pm\!0.027^{(4)}$ & $6.61\!\pm\!0.031^{(4)}$ & $\mathbf{1.73}\!\pm\!0.055^{(1)}$ & $\mathbf{1.73}\!\pm\!0.056^{(1)}$ & $\underline{6.60}\!\pm\!0.092^{(2)}$ & $\underline{6.51}\!\pm\!0.093^{(2)}$ & $1.39\!\pm\!0.010^{(4)}$ & $1.39\!\pm\!0.010^{(4)}$ \\
NeuralFlow & $7.41\!\pm\!0.122^{(8)}$ & $7.35\!\pm\!0.127^{(8)}$ & $3.46\!\pm\!0.234^{(10)}$ & $3.45\!\pm\!0.233^{(10)}$ & $6.77\!\pm\!0.064^{(9)}$ & $6.68\!\pm\!0.064^{(9)}$ & $3.11\!\pm\!0.305^{(11)}$ & $3.11\!\pm\!0.304^{(11)}$ \\
SeFT & $7.44\!\pm\!0.118^{(9)}$ & $7.36\!\pm\!0.115^{(9)}$ & $4.33\!\pm\!0.828^{(11)}$ & $4.32\!\pm\!0.827^{(11)}$ & $6.72\!\pm\!0.062^{(6)}$ & $6.61\!\pm\!0.059^{(5)}$ & $2.12\!\pm\!0.041^{(10)}$ & $2.12\!\pm\!0.041^{(10)}$ \\
Warpformer & $6.63\!\pm\!0.054^{(3)}$ & $6.56\!\pm\!0.059^{(3)}$ & $2.20\!\pm\!0.053^{(7)}$ & $2.19\!\pm\!0.054^{(7)}$ & $7.48\!\pm\!0.110^{(11)}$ & $7.35\!\pm\!0.109^{(11)}$ & $1.41\!\pm\!0.003^{(5)}$ & $1.41\!\pm\!0.003^{(5)}$ \\
mTAN & $7.61\!\pm\!0.010^{(11)}$ & $7.53\!\pm\!0.016^{(11)}$ & $2.96\!\pm\!0.314^{(9)}$ & $2.94\!\pm\!0.314^{(9)}$ & $6.70\!\pm\!0.062^{(5)}$ & $6.63\!\pm\!0.063^{(6)}$ & $1.85\!\pm\!0.018^{(8)}$ & $1.85\!\pm\!0.018^{(8)}$ \\
tPatchGNN & $6.75\!\pm\!0.070^{(6)}$ & $6.70\!\pm\!0.082^{(6)}$ & $2.16\!\pm\!0.247^{(5)}$ & $2.16\!\pm\!0.246^{(5)}$ & $6.73\!\pm\!0.083^{(7)}$ & $6.64\!\pm\!0.083^{(7)}$ & $1.45\!\pm\!0.035^{(7)}$ & $1.44\!\pm\!0.036^{(7)}$ \\
\bottomrule
\end{tabular}
\end{adjustbox}
\end{table*}

For fully synthetic data, the reference $R_{\mathrm{ct}}$ is
computed by trapezoidal integration over a dense common time
grid. For semi-synthetic data, it is computed as the equally
weighted squared error over all 512 future grid points.

\subsection{Real-World Datasets}
\label{app:real_datasets}

We evaluate models on eight real-world datasets spanning
healthcare, climate science, human sensing, finance, software
engineering, international events, and network systems.
These datasets exhibit diverse irregularity mechanisms,
including event-triggered logging, activity-dependent
collection, operational constraints, human scheduling, missing
observations, and system jitter~\cite{chang2025timeimm}. Only
the numerical time-series modalities are used.

The processed data are represented either as wide tables with
one row per record and timestamp or as sparse triplets of
timestamp, variable index, and value. Sparse observations are
converted into multivariate tensors with binary masks. Filled
values at unobserved positions are used only for tensor storage
and do not contribute to training or evaluation. Record
identifiers are deterministically divided into training,
validation, and test partitions, and forecasting windows are
constructed without crossing record boundaries.

\paragraph{CESNET.}
CESNET contains network traffic measurements collected from
11 devices, with 10 traffic variables. Its irregular timestamps
mainly arise from scheduling delay and logging jitter. Raw
timestamps are aggregated to minute resolution, and duplicate
observations within the same minute are averaged. We use
1,440 minutes of history to predict the following 1,440 minutes,
producing 252, 84, and 126 training, validation, and test
windows, respectively.

\paragraph{FNSPID.}
FNSPID contains six financial variables for 10 entities,
including open, high, low, close, adjusted close, and trading
volume. Observations are restricted by market operating hours,
creating structured gaps between trading periods. Timestamps
are converted into relative calendar days without filling
non-trading days. Both the historical and forecasting windows
contain 30 days, resulting in 496/171/176 train/validation/test
samples.

\paragraph{GDELT.}
GDELT records international events and contains five numerical
variables for eight event streams. Since observations are
generated when geopolitical events occur, the sampling process
is naturally event-driven. Individual events and fractional-day
timestamps are retained without daily aggregation. We use
14-day historical and forecasting windows, yielding
208/52/156 train/validation/test samples.

\paragraph{HumanActivity.}
HumanActivity contains 12 location variables obtained from
sensors placed on the body during different human activities.
The dataset contains 25 records, and the sensors are observed
at asynchronous and slightly different timestamps. Relative
timestamps are quantized, while duplicate measurements from
the same sensor and timestamp are averaged. The history and
forecasting lengths are 3,000 and 1,000 time units, producing
735/264/361 train/validation/test samples.

\paragraph{MIMIC-III.}
MIMIC-III is a large critical-care database containing
de-identified clinical records from ICU patients
~\cite{johnson2016mimic}. We combine laboratory, input,
output, and prescription events from the first 48 hours of each
admission and aggregate them into half-hour bins. The processed
data contain 22,406 admissions and 96 variables. The first
72 bins are used as history and the following 24 bins as the
forecasting horizon, resulting in 11,156/3,699/3,703
train/validation/test samples. Its irregularity mainly reflects
clinician-driven measurements and heterogeneous clinical
recording frequencies.

\begin{table*}[t]
\centering
\scriptsize
\setlength{\tabcolsep}{1.8pt}
\renewcommand{\arraystretch}{1.15}
\caption{Complete MAE and CAE results on MIMIC-III, RepoHealth, StudentLife, and USHCN. Values are mean $\pm$ standard deviation over three random seeds; standard deviations are rounded to three decimal places.}
\label{tab:app_mae_cae_part2}
\begin{adjustbox}{max width=\textwidth}
\begin{tabular}{l*{4}{cc}}
\toprule
Model & \multicolumn{2}{c}{\shortstack{MIMIC-III\\$\times10^{-1}$}} & \multicolumn{2}{c}{\shortstack{RepoHealth\\$\times10^{-1}$}} & \multicolumn{2}{c}{\shortstack{StudentLife\\$\times10^{-1}$}} & \multicolumn{2}{c}{\shortstack{USHCN\\$\times10^{-1}$}} \\
\cmidrule(lr){2-3}\cmidrule(lr){4-5}\cmidrule(lr){6-7}\cmidrule(lr){8-9}
& MAE & CAE & MAE & CAE & MAE & CAE & MAE & CAE \\
\midrule
APN & $4.78\!\pm\!0.034^{(6)}$ & $4.79\!\pm\!0.034^{(6)}$ & $\underline{3.47}\!\pm\!0.049^{(2)}$ & $\underline{3.55}\!\pm\!0.051^{(2)}$ & $\underline{5.75}\!\pm\!0.017^{(2)}$ & $5.78\!\pm\!0.017^{(3)}$ & $3.56\!\pm\!0.235^{(4)}$ & $3.70\!\pm\!0.291^{(5)}$ \\
ASTGI & $6.74\!\pm\!0.001^{(11)}$ & $6.76\!\pm\!0.001^{(11)}$ & $5.27\!\pm\!1.554^{(7)}$ & $5.36\!\pm\!1.584^{(7)}$ & $5.77\!\pm\!0.075^{(3)}$ & $\underline{5.77}\!\pm\!0.075^{(2)}$ & $3.11\!\pm\!0.080^{(3)}$ & $3.21\!\pm\!0.085^{(3)}$ \\
GRU-D & $4.98\!\pm\!0.018^{(7)}$ & $5.00\!\pm\!0.020^{(7)}$ & $7.56\!\pm\!0.614^{(11)}$ & $7.65\!\pm\!0.600^{(11)}$ & $6.15\!\pm\!0.023^{(8)}$ & $6.16\!\pm\!0.024^{(8)}$ & $4.53\!\pm\!0.408^{(11)}$ & $4.69\!\pm\!0.408^{(11)}$ \\
GraFITi & $\mathbf{4.40}\!\pm\!0.072^{(1)}$ & $\underline{4.47}\!\pm\!0.073^{(2)}$ & $\mathbf{3.21}\!\pm\!0.588^{(1)}$ & $\mathbf{3.26}\!\pm\!0.587^{(1)}$ & $5.83\!\pm\!0.035^{(6)}$ & $5.79\!\pm\!0.034^{(4)}$ & $\mathbf{2.97}\!\pm\!0.177^{(1)}$ & $\mathbf{3.08}\!\pm\!0.191^{(1)}$ \\
HyperIMTS & $\underline{4.46}\!\pm\!0.035^{(2)}$ & $\mathbf{4.41}\!\pm\!0.035^{(1)}$ & $3.89\!\pm\!0.370^{(3)}$ & $3.97\!\pm\!0.374^{(3)}$ & $\mathbf{5.74}\!\pm\!0.024^{(1)}$ & $\mathbf{5.76}\!\pm\!0.024^{(1)}$ & $4.15\!\pm\!0.967^{(8)}$ & $4.29\!\pm\!0.996^{(8)}$ \\
KAFNet & $4.60\!\pm\!0.035^{(3)}$ & $4.61\!\pm\!0.033^{(3)}$ & $5.17\!\pm\!1.978^{(6)}$ & $5.25\!\pm\!1.997^{(6)}$ & $5.78\!\pm\!0.023^{(4)}$ & $5.83\!\pm\!0.023^{(5)}$ & $4.31\!\pm\!0.291^{(9)}$ & $4.47\!\pm\!0.288^{(9)}$ \\
NeuralFlow & $5.11\!\pm\!0.015^{(8)}$ & $5.12\!\pm\!0.015^{(8)}$ & $7.36\!\pm\!0.510^{(10)}$ & $7.46\!\pm\!0.531^{(10)}$ & $6.49\!\pm\!0.075^{(10)}$ & $6.51\!\pm\!0.073^{(10)}$ & $3.58\!\pm\!0.615^{(5)}$ & $3.77\!\pm\!0.638^{(6)}$ \\
SeFT & $5.15\!\pm\!0.048^{(9)}$ & $5.16\!\pm\!0.046^{(9)}$ & $6.63\!\pm\!0.152^{(8)}$ & $6.72\!\pm\!0.152^{(8)}$ & $6.93\!\pm\!0.851^{(11)}$ & $6.94\!\pm\!0.849^{(11)}$ & $4.43\!\pm\!0.008^{(10)}$ & $4.59\!\pm\!0.004^{(10)}$ \\
Warpformer & $4.63\!\pm\!0.025^{(4)}$ & $4.64\!\pm\!0.023^{(4)}$ & $4.86\!\pm\!0.257^{(5)}$ & $4.98\!\pm\!0.260^{(5)}$ & $5.85\!\pm\!0.052^{(7)}$ & $5.86\!\pm\!0.052^{(7)}$ & $3.66\!\pm\!0.026^{(6)}$ & $3.67\!\pm\!0.022^{(4)}$ \\
mTAN & $5.38\!\pm\!0.043^{(10)}$ & $5.39\!\pm\!0.045^{(10)}$ & $6.96\!\pm\!0.242^{(9)}$ & $7.05\!\pm\!0.229^{(9)}$ & $6.19\!\pm\!0.127^{(9)}$ & $6.20\!\pm\!0.125^{(9)}$ & $\underline{3.01}\!\pm\!0.106^{(2)}$ & $\underline{3.13}\!\pm\!0.107^{(2)}$ \\
tPatchGNN & $4.65\!\pm\!0.023^{(5)}$ & $4.65\!\pm\!0.021^{(5)}$ & $4.36\!\pm\!0.322^{(4)}$ & $4.43\!\pm\!0.324^{(4)}$ & $5.82\!\pm\!0.081^{(5)}$ & $5.84\!\pm\!0.081^{(6)}$ & $3.82\!\pm\!0.725^{(7)}$ & $3.94\!\pm\!0.738^{(7)}$ \\
\bottomrule
\end{tabular}
\end{adjustbox}
\end{table*}

\paragraph{RepoHealth.}
RepoHealth describes issue, pull-request, and development
activities from four software repositories using 10 numerical
variables. Observations become denser during active development
periods and sparser during inactive periods. Timestamps are
converted into relative days, and duplicated daily observations
are averaged. We use 30-day history and forecasting windows,
with 97/59/50 train/validation/test samples.

\paragraph{StudentLife.}
StudentLife contains daily mobile-sensing measurements from
20 students, including activity, location, sleep, and phone-use
variables. Its nine variables are irregularly observed because
data availability depends on individual routines and device
usage. Both the historical and forecasting windows contain
30 days, producing 271/89/90 train/validation/test samples.

\paragraph{USHCN.}
USHCN contains long-term climate observations from 1,114
weather stations across the United States
~\cite{menne2015ushcn}. Five meteorological variables are used,
and missing daily observations make the resulting trajectories
sporadic. Following the commonly used irregular forecasting
protocol, the first 150 time units are used as history and the
following 50 units as the forecasting horizon. The resulting
split contains 666/220/224 train/validation/test samples.

\subsection{Baselines and Hyperparameters}
\label{app:baselines}

We compare eleven representative models covering recurrent,
continuous-time, set-based, attention-based, graph-based, and
pre-alignment forecasting paradigms. To ensure consistent
comparisons, each model uses the same configuration across all
real-world datasets.

\paragraph{APN.}
APN aggregates irregular observations into learnable temporal
patches through soft boundaries and predicts values at arbitrary
query timestamps~\cite{liu2026apn}. We use hidden dimension
$d=56$, two layers, eight attention heads, time-embedding
dimension 8, two patches, dropout 0.1, and batch size 32.

\paragraph{ASTGI.}
ASTGI directly represents each discrete observation as a point
in a learnable spatio-temporal space and adaptively constructs
causal nearest-neighbor graphs for information propagation
~\cite{liu2026astgi}. We use hidden dimension 128, two
propagation layers, time-embedding dimension 128,
channel-embedding dimension 64, 96 candidate neighbors, MLP
ratio 4, dropout 0.1, and batch size 8.

\paragraph{GRU-D.}
GRU-D extends recurrent models with trainable decay mechanisms
that explicitly incorporate missingness masks and elapsed time
between observations~\cite{che2016grud}. We use one
recurrent layer with hidden dimension 128 and batch size 32.

\paragraph{GraFITi.}
GraFITi represents observed and queried values as nodes in a
bipartite graph and performs graph-based information propagation
for irregular forecasting~\cite{yalavarthi2024grafiti}. We use
hidden dimension 128, four graph layers, one attention head,
zero dropout, and batch size 32.

\paragraph{HyperIMTS.}
HyperIMTS constructs hypergraph interactions to model
higher-order dependencies among asynchronously observed
variables~\cite{li2025hyperimts}. We use hidden dimension 128,
three layers, one attention head, dropout 0.1, and batch size 32.

\paragraph{KAFNet.}
KAFNet employs kernel aggregation to pre-align irregular
observations before temporal modeling~\cite{zhou2026kafnet}.
We use hidden dimension 16, pre-convolution dimension 32, one
layer, one attention head, four Gaussian kernels,
time-embedding dimension 10, and batch size 32.

\paragraph{NeuralFlow.}
NeuralFlow parameterizes continuous latent dynamics through an
invertible flow, providing an efficient alternative to numerical
ODE solvers~\cite{bilos2021neuralflows}. We use hidden
dimension 100, two coupling-flow layers, latent dimension 20,
time-encoding hidden dimension 8, three hidden layers,
reconstruction dimension 30, and batch size 32.

\paragraph{SeFT.}
SeFT treats all irregular observations as an unordered set and
uses learned set functions and attention pooling to obtain a
fixed-dimensional representation~\cite{horn2020seft}. We use
hidden dimension 128, three $\phi/\rho$ layers, two $\psi$
layers, four attention heads, positional dimension 4,
dropout 0.1, and batch size 32.

\begin{figure*}[t]
\centering
\begin{minipage}[t]{0.49\textwidth}
    \centering
    \includegraphics[width=\linewidth]
    {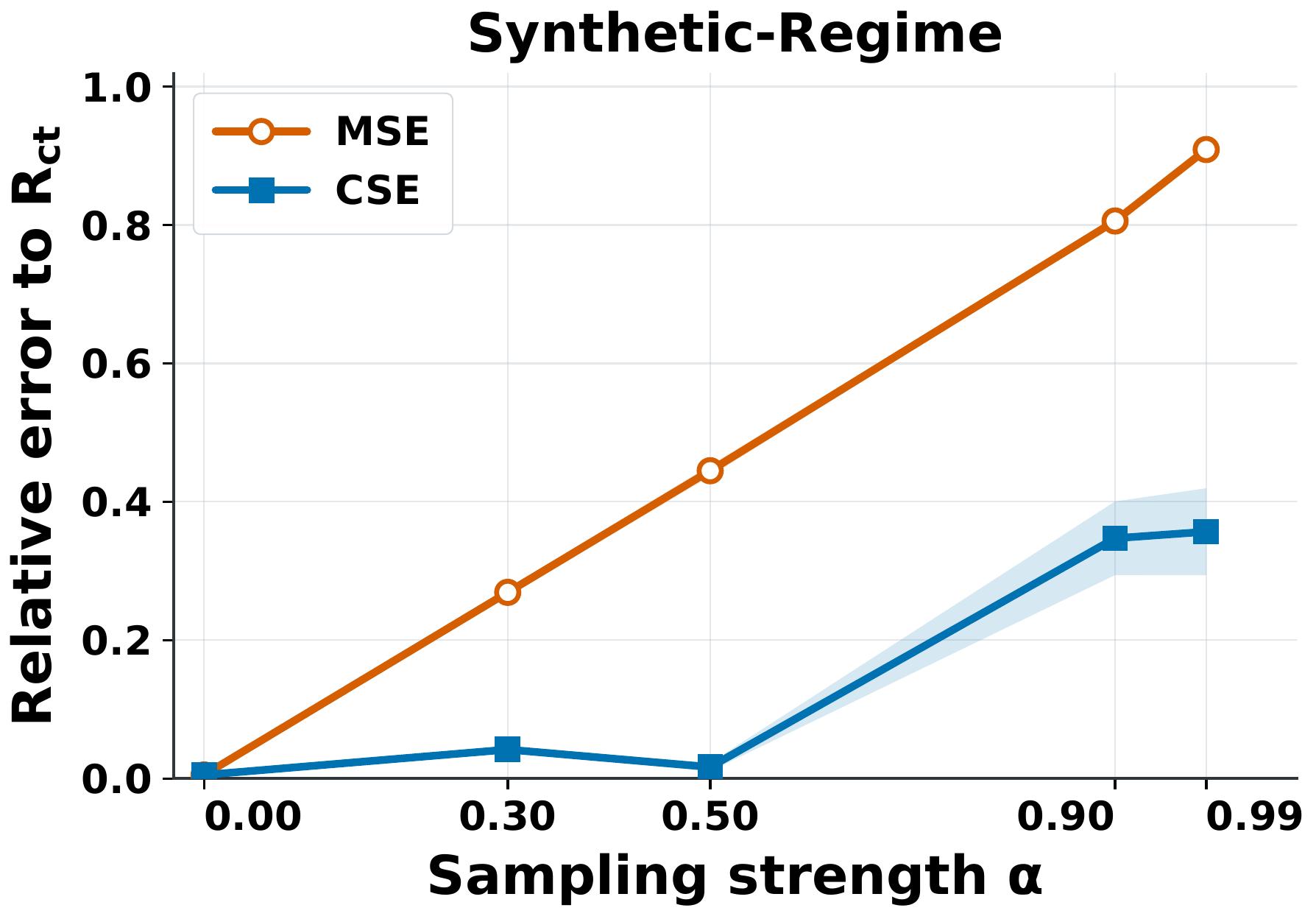}\\[-2pt]
    \small (a) Synthetic-Regime
\end{minipage}
\hfill
\begin{minipage}[t]{0.49\textwidth}
    \centering
    \includegraphics[width=\linewidth]
    {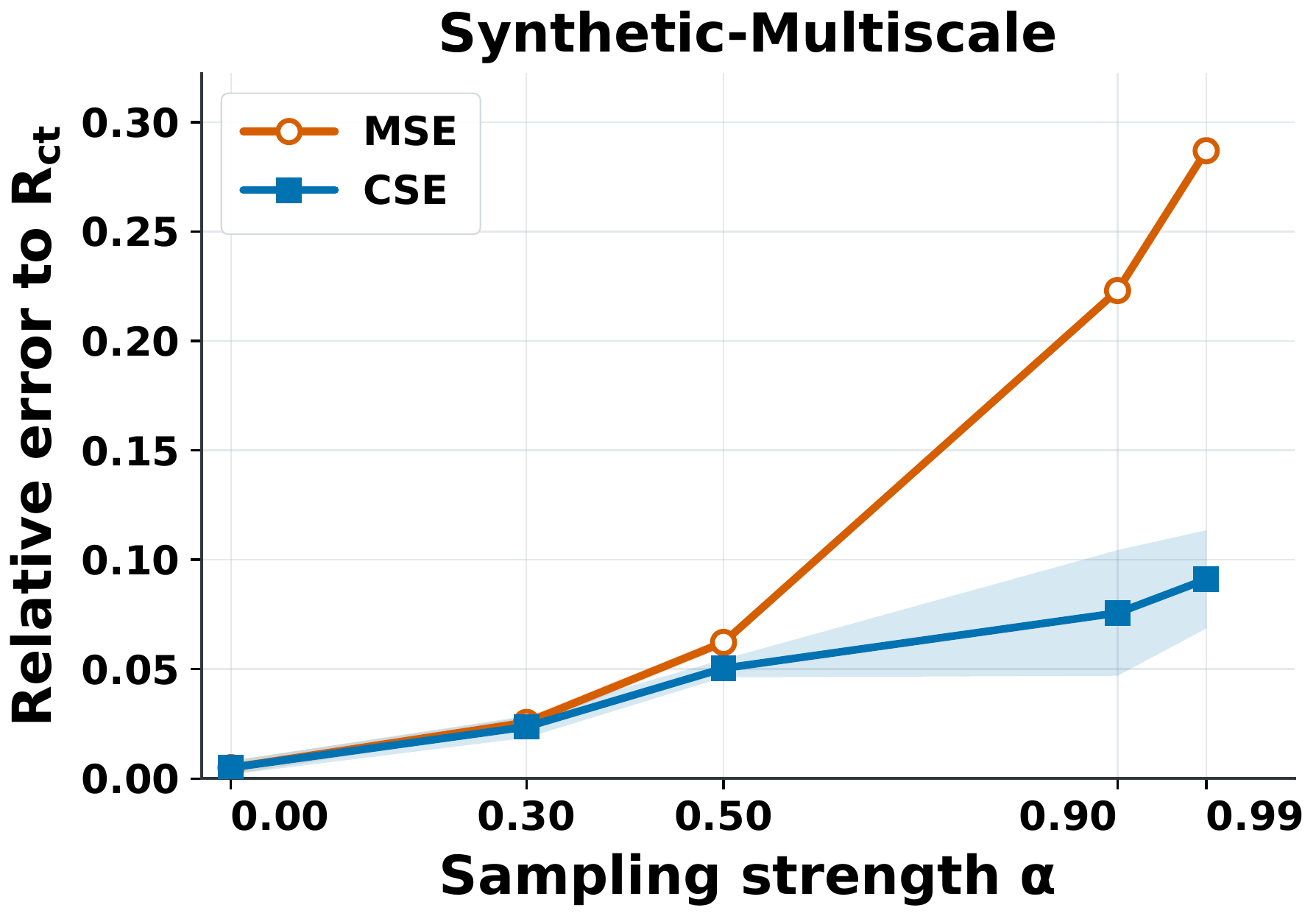}\\[-2pt]
    \small (b) Synthetic-Multiscale
\end{minipage}

\vspace{4pt}

\begin{minipage}[t]{0.49\textwidth}
    \centering
    \includegraphics[width=\linewidth]
    {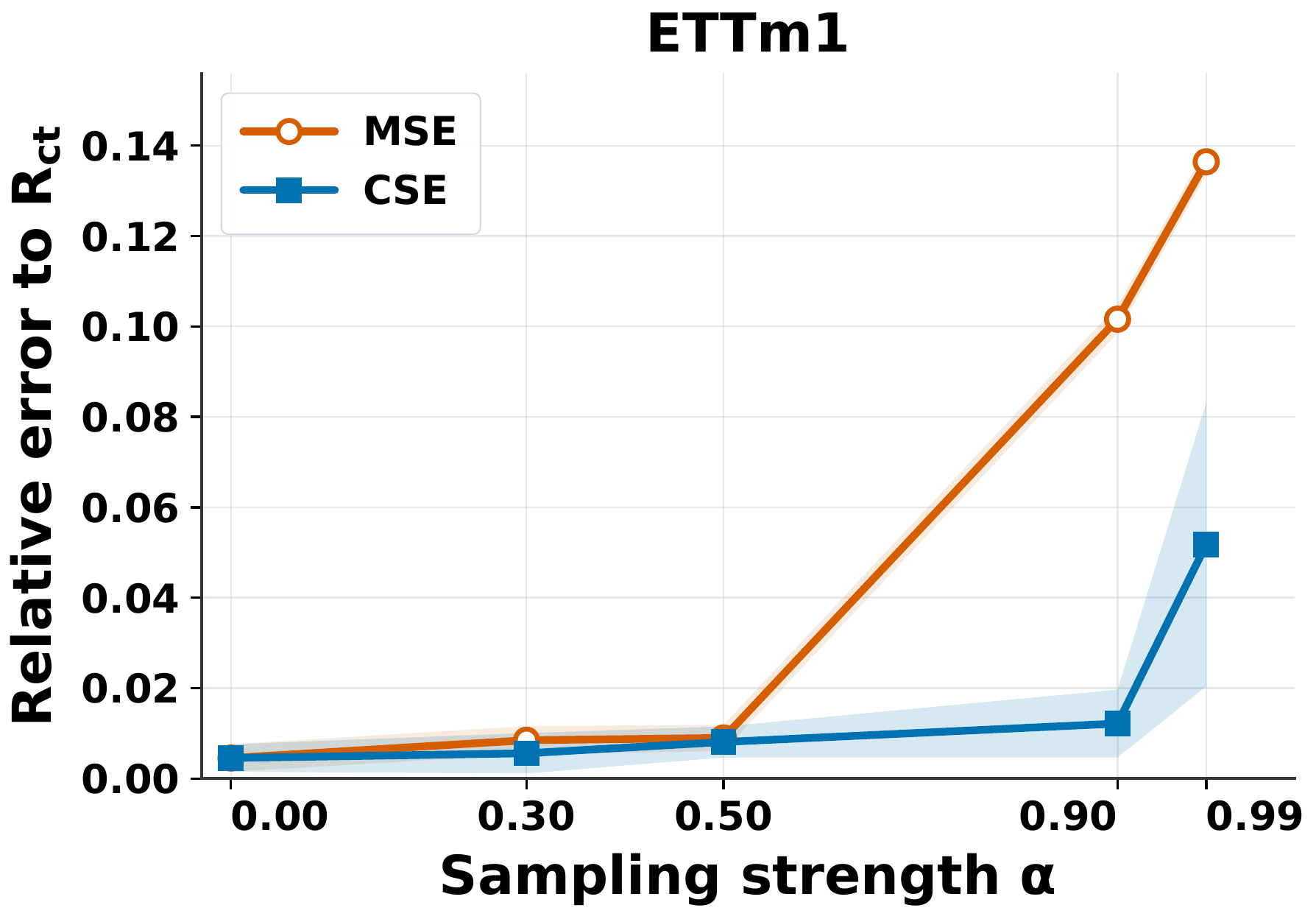}\\[-2pt]
    \small (c) ETTm1
\end{minipage}
\hfill
\begin{minipage}[t]{0.49\textwidth}
    \centering
    \includegraphics[width=\linewidth]
    {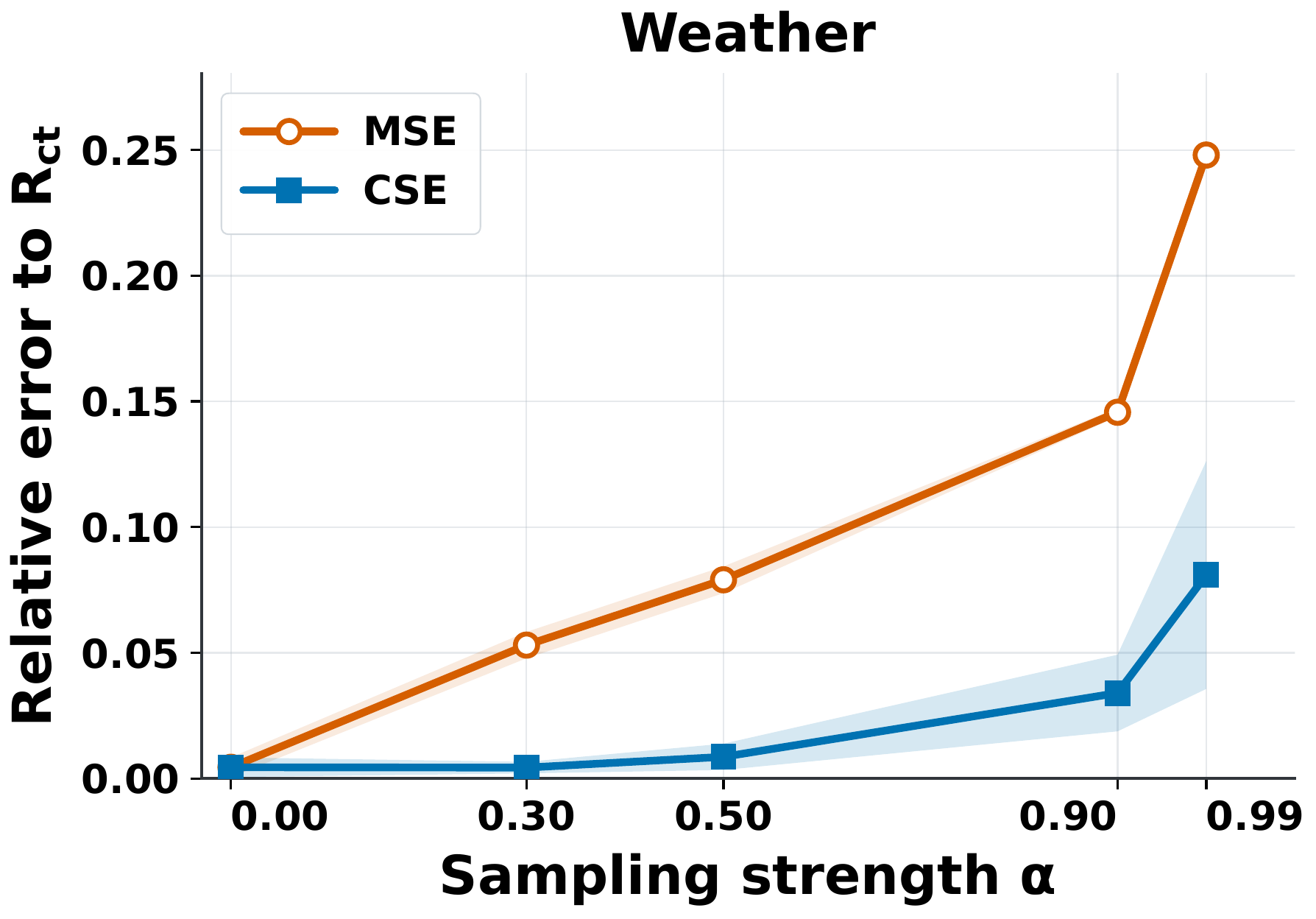}\\[-2pt]
    \small (d) Weather
\end{minipage}
\caption{Relative estimation errors of MSE and CSE with
respect to $R_{\mathrm{ct}}$ under increasing sampling
non-uniformity on all synthetic and semi-synthetic datasets.}
\label{fig:app_risk_recovery_all}
\end{figure*}

\paragraph{Warpformer.}
Warpformer applies multiscale temporal warping to transform
irregular observations into representations suitable for
Transformer-based modeling~\cite{zhang2023warpformer}. We use
hidden dimension 128, three layers, inner dimension 64,
one attention head, key/value dimensions of 8, and batch size 32.

\paragraph{mTAN.}
mTAN employs multi-time attention to map irregular observations
onto continuous reference timestamps
~\cite{shukla2021mtan}. We use hidden dimension 128, latent
dimension 32, one attention head, 128 reference points,
observation standard deviation 0.01, and batch size 32.

\paragraph{tPatchGNN.}
tPatchGNN organizes irregular observations into transformable
temporal patches and models variable interactions using graph
neural networks~\cite{zhang2024tpatchgnn}. We use hidden
dimension 32, one graph layer, one attention head, one
Transformer layer, one-hop propagation, node-embedding
dimension 10, time-embedding dimension 10, a linear output
head, and batch size 32.

\paragraph{Controlled-experiment baselines.}
The controlled experiments use SeFT, CRU, GraFITi, tPatchGNN,
and KAFNet. SeFT uses hidden dimension 64, two $\phi/\rho$
layers, two $\psi$ layers, two attention heads, and dropout
0.02. CRU uses hidden dimension 20 and 20 basis functions
~\cite{schirmer2022cru}. GraFITi uses hidden dimension 28 and
two graph layers. tPatchGNN uses hidden dimension 32, patch
length 4, one graph layer, and one Transformer layer. KAFNet
uses hidden dimension 64, pre-convolution dimension 32, two
layers, eight Gaussian kernels, and time-embedding dimension 16.

\begin{table*}[t]
\centering
\scriptsize
\setlength{\tabcolsep}{3.0pt}
\renewcommand{\arraystretch}{1.15}
\caption{Complete model-evaluation results on the two synthetic datasets under $\alpha=0.9$. Values are mean $\pm$ standard deviation over 20 timestamp resamples; $R_{\mathrm{ct}}$ is computed from the complete trajectory. Superscripts denote ranks.}
\label{tab:app_controlled_synthetic}
\begin{adjustbox}{max width=\textwidth}
\begin{tabular}{l*{2}{ccc}}
\toprule
Model & \multicolumn{3}{c}{\shortstack{Synthetic-Regime\\$\times10^{-2}$}} & \multicolumn{3}{c}{\shortstack{Synthetic-Multiscale\\$\times10^{-2}$}} \\
\cmidrule(lr){2-4}\cmidrule(lr){5-7}
& MSE & CSE & $R_{\mathrm{ct}}$ & MSE & CSE & $R_{\mathrm{ct}}$ \\
\midrule
SeFT & $1.0884\!\pm\!0.0228^{(3)}$ & $\mathbf{8.6994}\!\pm\!0.3760^{(1)}$ & $\mathbf{6.2552}^{(1)}$ & $1.2655\!\pm\!0.0058^{(3)}$ & $1.1169\!\pm\!0.0632^{(3)}$ & $1.1659^{(3)}$ \\
CRU & $\mathbf{0.6527}\!\pm\!0.0232^{(1)}$ & $\underline{8.8105}\!\pm\!0.3814^{(2)}$ & $\underline{6.4831}^{(2)}$ & $1.3118\!\pm\!0.0055^{(4)}$ & $\mathbf{0.8085}\!\pm\!0.0380^{(1)}$ & $\mathbf{0.9575}^{(1)}$ \\
GraFITi & $2.5741\!\pm\!0.0233^{(5)}$ & $9.8776\!\pm\!0.4644^{(5)}$ & $7.6588^{(5)}$ & $1.6822\!\pm\!0.0053^{(5)}$ & $1.4708\!\pm\!0.0477^{(5)}$ & $1.5059^{(5)}$ \\
tPatchGNN & $\underline{0.9132}\!\pm\!0.0241^{(2)}$ & $9.3798\!\pm\!0.3603^{(4)}$ & $6.9119^{(4)}$ & $\underline{0.8939}\!\pm\!0.0052^{(2)}$ & $1.1272\!\pm\!0.0687^{(4)}$ & $1.1789^{(4)}$ \\
KAFNet & $1.5435\!\pm\!0.0232^{(4)}$ & $9.0437\!\pm\!0.3602^{(3)}$ & $6.7512^{(3)}$ & $\mathbf{0.8088}\!\pm\!0.0057^{(1)}$ & $\underline{1.0751}\!\pm\!0.0511^{(2)}$ & $\underline{1.1567}^{(2)}$ \\
\bottomrule
\end{tabular}
\end{adjustbox}
\end{table*}

\subsection{Training and Evaluation Details}
\label{app:implementation}

\paragraph{Training protocol.}
All models are trained using the Adam optimizer and masked
observation-point MSE. For prediction mask $m_{bid}$, the
training objective is
\begin{equation}
\mathcal{L}_{\mathrm{train}}
=
\frac{
\sum_{b,i,d}
m_{bid}
\left(
\widehat y_{bid}-y_{bid}
\right)^2
}{
\sum_{b,i,d}m_{bid}
}.
\label{eq:app_training_loss}
\end{equation}
All baselines are trained for 100 epochs, and the checkpoint
with the lowest validation MSE is retained for testing. The
learning rate is $10^{-3}$ for all models. NeuralFlow and mTAN
use weight decay $10^{-4}$, while the remaining real-world
baselines use no weight decay. Models that originally employ
probabilistic objectives, including mTAN and NeuralFlow, are
also optimized using the common MSE objective to ensure a
unified training protocol.

The controlled experiments use learning rate $10^{-3}$,
weight decay $10^{-5}$, batch size 64, and global gradient-norm
clipping at 1.0. Training and validation target timestamps are
uniformly sampled from the future interval.

All model--dataset experiments are independently repeated with
three random seeds, 2024, 2025, and 2026. We report the mean
and standard deviation across the three runs. In the controlled
sampling-shift experiment, each trained checkpoint is further
evaluated using 20 independent timestamp resamples for each
sampling strength $\alpha$.

\paragraph{Temporal-density estimation.}
Future timestamps are mapped to the common forecasting
interval $[0,1]$ before density estimation. We use a
trajectory-specific leave-one-out Gaussian kernel density
estimator. The bandwidth follows a robust Silverman rule:
\begin{equation}
h_s
=
\max
\left\{
0.9
\min\left(
\sigma_s,
\frac{\operatorname{IQR}_s}{1.34}
\right)
L_s^{-1/5},
\frac{1}{\max(20,2L_s)},
10^{-6}
\right\}.
\label{eq:app_bandwidth}
\end{equation}
Non-positive scale estimates are excluded when computing the
minimum. Estimated densities are lower-bounded by $10^{-8}$,
and no importance-weight clipping or boundary reflection is
applied. Conditional and marginal timestamp densities are
estimated only from test timestamps and are shared by all
models evaluated on the same dataset.

\paragraph{Metric computation.}
For each model--dataset--seed combination, the selected
checkpoint is loaded once, and MSE, GSE, and CSE are computed
from the same test predictions. Let $m_{sid}$ indicate whether
variable $d$ is observed at future timestamp $t_{s,i}$ and let
$w_{s,i}$ denote the corresponding temporal weight. The
trajectory-level weighted error is
\begin{equation}
\widehat R_s(w)
=
\frac{
\sum_{i,d}
m_{sid}w_{s,i}
\left(
\widehat y_{sid}-y_{sid}
\right)^2
}{
\sum_{i,d}m_{sid}w_{s,i}
}.
\label{eq:app_weighted_metric}
\end{equation}
The observation-distribution, global-time, and continuous-time
weights are
\begin{equation}
\begin{aligned}
w^{\mathrm{obs}}_{s,i}
&=1,\\
w^{\mathrm{global}}_{s,i}
&=
\frac{
\widehat p_T(t_{s,i})
}{
\widehat p(t_{s,i}\mid X_s)
},\\
w^{\mathrm{ct}}_{s,i}
&=
\frac{1}{
\widehat p(t_{s,i}\mid X_s)
}.
\end{aligned}
\label{eq:app_temporal_weights}
\end{equation}
Each trajectory is normalized independently, after which the
trajectory-level errors are averaged with equal weight.
Padding positions and timestamps without any observed target
variables are excluded from both density estimation and metric
computation.

\paragraph{Implementation environment.}
All experiments are implemented in Python 3.12 and PyTorch
2.10.0 with CUDA 12.8. Training and evaluation are conducted
on NVIDIA GeForce RTX 4090 GPUs.

\section{Additional Experimental Results}
\label{app:additional_results}

\subsection{Complete Results on Synthetic and Semi-Synthetic Data}
\label{app:controlled_results}

Figure~\ref{fig:app_risk_recovery_all} reports the relative
estimation errors of MSE and CSE with respect to
$R_{\mathrm{ct}}$ on all four controlled datasets. The curves
show the mean over 20 independent test-timestamp resamples,
and the shaded regions denote one standard deviation. Because
the trained models and underlying future trajectories remain
fixed across different values of $\alpha$, the observed changes
are caused solely by the shift in the test-time sampling
distribution.
\begin{table*}[t]
\centering
\scriptsize
\setlength{\tabcolsep}{3.0pt}
\renewcommand{\arraystretch}{1.15}
\caption{Complete model-evaluation results on the two semi-synthetic datasets under $\alpha=0.9$. Values are mean $\pm$ standard deviation over 20 timestamp resamples; $R_{\mathrm{ct}}$ is computed on the complete future grid. Superscripts denote ranks.}
\label{tab:app_controlled_semisynthetic}
\begin{adjustbox}{max width=\textwidth}
\begin{tabular}{l*{2}{ccc}}
\toprule
Model & \multicolumn{3}{c}{\shortstack{ETTm1\\$\times10^{0}$}} & \multicolumn{3}{c}{\shortstack{Weather\\$\times10^{-1}$}} \\
\cmidrule(lr){2-4}\cmidrule(lr){5-7}
& MSE & CSE & $R_{\mathrm{ct}}$ & MSE & CSE & $R_{\mathrm{ct}}$ \\
\midrule
SeFT & $1.1032\!\pm\!0.0059^{(4)}$ & $1.1145\!\pm\!0.0153^{(4)}$ & $1.1130^{(4)}$ & $4.4046\!\pm\!0.0146^{(5)}$ & $4.9629\!\pm\!0.0629^{(5)}$ & $4.8493^{(5)}$ \\
CRU & $1.3394\!\pm\!0.0058^{(5)}$ & $1.1001\!\pm\!0.0153^{(3)}$ & $1.0994^{(3)}$ & $\mathbf{3.0839}\!\pm\!0.0144^{(1)}$ & $4.6087\!\pm\!0.0667^{(3)}$ & $4.4631^{(3)}$ \\
GraFITi & $1.0886\!\pm\!0.0056^{(3)}$ & $1.4140\!\pm\!0.0163^{(5)}$ & $1.4016^{(5)}$ & $3.8918\!\pm\!0.0147^{(4)}$ & $4.9221\!\pm\!0.0665^{(4)}$ & $4.7612^{(4)}$ \\
tPatchGNN & $\underline{0.9251}\!\pm\!0.0056^{(2)}$ & $\mathbf{0.9330}\!\pm\!0.0151^{(1)}$ & $\mathbf{0.9302}^{(1)}$ & $3.8789\!\pm\!0.0181^{(3)}$ & $\mathbf{3.6856}\!\pm\!0.0645^{(1)}$ & $\mathbf{3.5274}^{(1)}$ \\
KAFNet & $\mathbf{0.9093}\!\pm\!0.0056^{(1)}$ & $\underline{0.9640}\!\pm\!0.0155^{(2)}$ & $\underline{0.9584}^{(2)}$ & $\underline{3.4839}\!\pm\!0.0144^{(2)}$ & $\underline{3.7783}\!\pm\!0.0607^{(2)}$ & $\underline{3.6496}^{(2)}$ \\
\bottomrule
\end{tabular}
\end{adjustbox}
\end{table*}
Under uniform sampling, the inverse-density weights are
constant and CSE consequently reduces to MSE. As $\alpha$
increases, observations become increasingly concentrated in a
limited portion of the future interval. MSE therefore places
disproportionate emphasis on densely sampled regions, and its
deviation from $R_{\mathrm{ct}}$ generally increases. In
contrast, CSE compensates for this concentration through
inverse-density weighting and remains substantially closer to
$R_{\mathrm{ct}}$ under strong non-uniform sampling. Although
density estimation and importance weighting may introduce
additional finite-sample variability, the correction becomes
increasingly beneficial as the mismatch between the observed
and uniform temporal distributions grows. The consistent
behavior on ETTm1 and Weather further shows that this
advantage is not restricted to analytically generated
trajectories.

Tables~\ref{tab:app_controlled_synthetic} and
\ref{tab:app_controlled_semisynthetic} report the complete
model-level results under $\alpha=0.9$. On all four datasets,
CSE recovers exactly the same ranking of the five models as
$R_{\mathrm{ct}}$, whereas MSE produces at least one ranking
inversion. This result indicates that the effect of temporal
sampling is not limited to a uniform shift in metric values:
different models are affected to different degrees because
their prediction errors vary differently over time. Correcting
the target-time distribution therefore improves both risk
estimation and the recovery of model comparisons, supporting
the main-text conclusion that CSE more accurately reflects
continuous-time predictive performance under non-uniform
sampling.

\subsection{Complete Results on Real-World Datasets}
\label{app:real_complete_results}

Tables~\ref{tab:app_mse_cse_part1} and
\ref{tab:app_mse_cse_part2} report the complete MSE and CSE
results on the eight real-world datasets, including standard
deviations over three random seeds. Tables~\ref{tab:app_mae_cae_part1}
and \ref{tab:app_mae_cae_part2} provide the corresponding MAE
and Continuous-Time Absolute Error (CAE) results. The best and
second-best mean values are shown in bold and underlined,
respectively, and superscripts denote ranks computed from the
unrounded means.

Consistent with the observations in the main text, CSE is
neither a fixed-direction nor a fixed-ratio transformation of
MSE. On CESNET, FNSPID, GDELT, and HumanActivity, CSE is
generally lower than MSE, whereas the opposite trend is more
common on MIMIC-III, RepoHealth, and USHCN. Moreover, even
within the same dataset, the magnitude of the change varies
across models. This confirms that the sampling distribution
interacts with model-specific temporal error patterns rather
than simply applying a dataset-level scaling factor. The
reported standard deviations also distinguish this systematic
metric effect from training variability: most model--dataset
combinations remain relatively stable across seeds, although
datasets such as RepoHealth and USHCN exhibit larger variation
for several models.

The MAE--CAE comparison exhibits a similar pattern. The
direction of the correction is broadly consistent with the
MSE--CSE comparison, while its magnitude remains dependent on
both the dataset and the model. In particular, the
best-performing model on MIMIC-III changes from GraFITi under
MAE to HyperIMTS under CAE, accompanied by additional local
rank changes on several datasets. Therefore, the discrepancies
reported in the main text are not specific to squared error,
but arise more generally from the temporal distribution under
which pointwise prediction losses are aggregated. Since the
complete continuous trajectories of real-world datasets are
unavailable, these results do not directly identify which
metric is exact; rather, the systematic differences suggest
that relying solely on observation-point metrics may not fully
characterize models' continuous-time predictive performance.

\end{document}

%% file: paper/introduction.tex
\section{Introduction}
Irregular time series are ubiquitous in real-world applications such as healthcare, environmental monitoring, and human activity analysis~\cite{DBLP:conf/aaai/GhassemiPNBCSF15,DBLP:journals/sensors/DecorteMLMLMV24,afrifa2020missing}. 
Unlike regularly sampled time series, their observations occur at non-uniform timestamps, and different samples may exhibit different sampling patterns. 
In recent years, considerable research effort has been devoted to modeling such non-uniform temporal processes,  
leading to the development of continuous-time~\cite{bilos2021neuralflows}, attention-based~\cite{shukla2021mtan}, graph-based~\cite{yalavarthi2024grafiti}, and patch-based~\cite{liu2026apn} models.
However, compared with the rapid progress in model design, evaluation metrics for irregular time-series forecasting remain insufficiently studied.

\begin{figure}
    \centering
    \includegraphics[width=\linewidth]{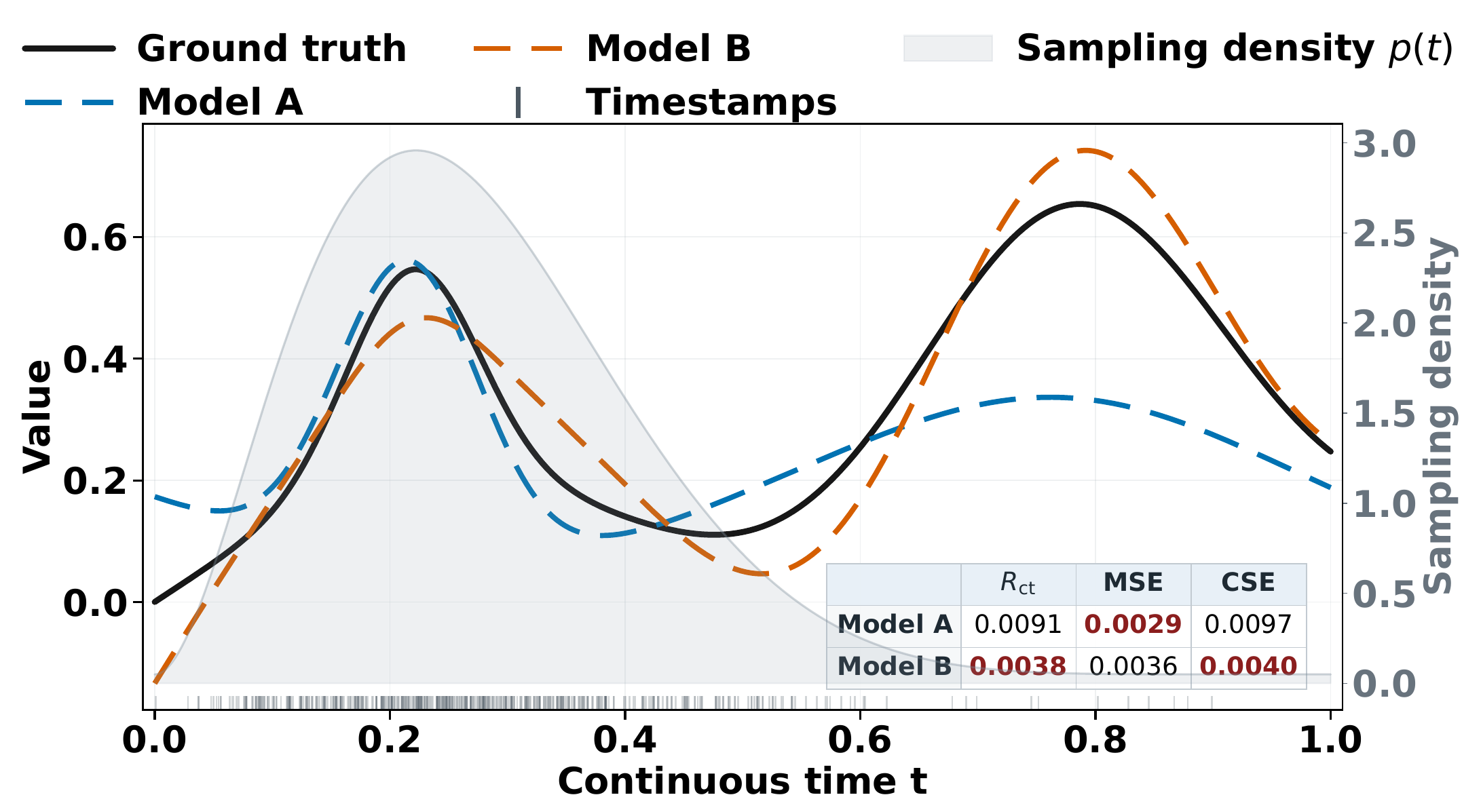}
    \caption{Illustration of evaluation bias under non-uniform sampling. Although Model B has lower continuous-time error, Model A obtains a lower MSE because of over-dense sampling within the interval $[0.0, 0.4]$.}
    \label{fig:intro}
\end{figure}

For time-series forecasting, an ideal evaluation metric is the continuous-time discrepancy between the predicted trajectory and the true trajectory $X$ over the entire future interval. 
Given $X$ over a time span $\mathcal{T}$, let $\ell(t,X)$ denote the prediction error of a model at time $t$. This continuous-time discrepancy is defined as the \textit{continuous-time risk}:
\begin{equation}
R_{\mathrm{ct}}
=
\mathbb{E}_X
\left[
\frac{1}{|\mathcal{T}|}
\int_{\mathcal{T}}
\ell(t,X)\,dt
\right].
\label{eq:intro_ct_risk}
\end{equation}
This risk directly characterizes the model's continuous-time predictive performance over the future interval.

Existing studies typically adopt mean squared error (MSE) and mean absolute error (MAE), which are widely used in regularly sampled forecasting. 
We argue that \textbf{in irregular settings, MSE and MAE can be affected by the sample-specific timestamp sampling distributions and may therefore fail to accurately reflect a model's performance.} 
As illustrated in Fig.~\ref{fig:intro}, although Model B has a lower continuous-time error than Model A  (i.e., the orange line of Model B is closer to the black ground truth line over the whole time span), Model A obtains a lower MSE because of over-dense sampling within the interval $[0.0, 0.4]$. 
By theoretical analysis, we find that MSE provides a biased estimate of the continuous-time risk $R_{ct}$, as it actually measures the observation-time risk $R_{\mathrm{obs}}=\mathbb{E}_X[\int_{\mathcal{T}}\ell(t,X)p(t\mid X)\,dt]$, 
where $p(t\mid X)$ is the conditional sampling distribution for the timestamps.
This distribution is sample-specific and thus causes model performance evaluations to be confounded by sample-specific sampling patterns.

To eliminate the influence of the timestamp sampling distributions on evaluation results, we begin directly from $R_{\mathrm{ct}}$ and construct its empirical estimator, termed the \textit{Continuous-time Squared Error (CSE)}, through importance sampling. 
CSE reweights errors according to the observation-time density $p(t\mid X)$, reducing the contribution of densely sampled regions while increasing that of sparsely sampled regions, thereby mitigating evaluation bias caused by timestamp sampling distributions. 
Returning to Fig.~\ref{fig:intro}, CSE correctly identifies Model B as having the lower continuous-time error.
We further prove that the asymptotic estimation error of CSE with respect to the continuous-time risk is no greater than that of MSE. 
Finally, we further decompose the discrepancy between MSE and CSE into a sampling-dependence gap and a temporal-distribution gap to investigate their respective effects in real-world settings.
Similarly, we can derive the \textit{Continuous-time Absolute Error (CAE)} as the debiased counterpart to MAE.

To evaluate the effectiveness of CSE, we require access to the true trajectory $X$ of each dataset. To this end, we construct a systematic benchmark for irregular time-series forecasting covering synthetic, semi-synthetic, and real-world datasets. Both the synthetic and semi-synthetic datasets have the true continuous or regular trajectory. Based on a comprehensive evaluation of this benchmark, we have the following conclusions:
(1) on synthetic and semi-synthetic datasets, CSE can recover continuous-time risk more accurately than MSE under non-uniform sampling;
(2) on real-world datasets, relying solely on MSE may not fully reflect models' continuous-time predictive performance;
(3) the discrepancies between MSE and CSE are jointly influenced by the sampling-dependence and temporal-distribution gaps.

The main contributions are summarized as follows:
\begin{itemize}
    \item \textbf{Revealing the biased evaluation of MSE for model's continuous-time predictive performance.}
    We show that MSE estimates the observation-distribution risk rather than the continuous-time risk, thereby coupling model evaluation with sampling mechanisms.

    \item \textbf{A continuous-time evaluation metric with theoretical guarantees.}
    We propose Continuous-Time Squared Error (CSE), which estimates the continuous-time risk from irregular timestamps through importance weighting, and prove that its asymptotic estimation error is no greater than MSE.

    \item \textbf{A systematic benchmark for continuous-time evaluation.}
    We construct a benchmark covering synthetic, semi-synthetic, and real-world datasets to validate the effectiveness of CSE, systematically evaluate models' continuous-time predictive performance, and analyze the sources of discrepancies between MSE and CSE.
\end{itemize}

%% file: paper/related_work.tex
\section{Related Work}

\subsection{Irregular Time Series Forecasting}
Irregular time series are typically characterized by sparse, asynchronous, and nonuniform sampling. Existing studies have primarily focused on modeling continuous-time dynamics from limited observations. One line of work explicitly describes the continuous evolution of latent states. For example, Latent ODE, Neural CDE, and Continuous Recurrent Unit handle arbitrary time intervals based on ordinary differential equations, controlled differential equations, and continuous state transitions, respectively~\cite{rubanova2019latent,kidger2020neuralcde,schirmer2022cru}. Another line of work directly models irregular observations together with their temporal information. SeFT represents a time series as a set of observations, mTAN employs multi-time attention to aggregate asynchronous observations, and Raindrop uses graph structures to capture dynamic dependencies among variables~\cite{horn2020seft,shukla2021mtan,zhang2022raindrop}. More recent methods further integrate multiscale representations with graph structures. For instance, Warpformer models temporal patterns at different scales through time warping, whereas GraFITi formulates irregular forecasting as an information propagation process over a graph~\cite{zhang2023warpformer,yalavarthi2024grafiti}.

Despite these advances, existing studies primarily focus on model design and typically evaluate predictions using MSE or MAE at observed future timestamps. In contrast, we study the evaluation protocol itself and examine how irregular sampling affects model comparison and selection.

\subsection{Benchmarks and Evaluation for Irregular Time Series}
With the growing interest in irregular time series, prior studies have developed related benchmarks from the perspectives of data resources, software frameworks, and standardized experimental settings. The MIMIC-IV benchmark organizes unified tasks for sparse and irregular clinical time series and compares a range of representative models~\cite{bui2024benchmarking}. PYRREGULAR provides unified data structures and analytical tools, together with a standardized benchmark covering multiple datasets and classification methods~\cite{spinnato2025pyrregular}. Time-IMM further introduces new datasets and evaluation benchmarks for irregular multimodal multivariate time series~\cite{chang2025timeimm}. For forecasting tasks, Physiome-ODE constructs controlled datasets from biological ODEs to provide more challenging and discriminative environments for comparing irregular forecasting models~\cite{klotergens2025physiome}.

Existing benchmarks mainly improve datasets, tasks, and experimental standardization rather than the evaluation objective itself. Physiome-ODE is the most closely related work: it provides controlled ODE-generated trajectories to improve model discrimination, but still evaluates predictions using MSE at sampled query times~\cite{klotergens2025physiome}. Consequently, it does not examine whether model conclusions depend on the query-time distribution or recover risk over the entire future interval. In contrast, we take continuous-time risk as the evaluation target, propose CSE to estimate it, and decompose the resulting evaluation discrepancy into sampling-dependence and temporal-distribution gaps.

%% file: paper/preliminary.tex
\section{Preliminaries}

\subsection{Irregular Time Series Forecasting}
The dataset contains $S$ irregular time series. For the $s$-th sequence, its historical observations are denoted by
$\mathcal{O}_s=\{(q_{s,i},\mathbf{x}_{s,i})\}_{i=1}^{H_s}$, where $q_{s,i}\in\mathbb{R}$ and $\mathbf{x}_{s,i}\in\mathbb{R}^D$
denote the irregular timestamp and the corresponding multivariate observation, respectively. Given the historical observations $\mathcal{O}_s$, a forecasting model $f_{\theta}$ aims to estimate the latent process over a future time interval $\mathcal{T}$. For any query time $t\in\mathcal{T}$, the model output is given by $\hat{\mathbf{X}}_s(t)=f_{\theta}(\mathcal{O}_s,t)$. Let $\mathbf{X}_s(t)$ denote the ground-truth latent process, and define the prediction error at time $t$ as $\ell(t,X_s)=\mathcal{L}(\mathbf{X}_s(t),\hat{\mathbf{X}}_s(t))$. In practical test datasets, the ground-truth values are available only at a finite set of future timestamps $\{t_{s,j}\}_{j=1}^{L_s}$, which are generated from the conditional sampling distribution $p(t\mid X_s)$.

\subsection{MSE and MAE under Irregular Sampling}
Existing studies on irregular time series forecasting typically compute MSE and MAE at the actually observed future timestamps.
In the following, we use MSE as the example to illustrate the idea, i.e.,
$\ell(t,X_s)=\left\|\mathbf{X}_s(t)-\hat{\mathbf{X}}_s(t)\right\|^2$.
Accordingly, the MSE on the test set can be written as
\begin{equation}
\mathrm{MSE}
=
\frac{1}{S}
\sum_{s=1}^{S}
\frac{1}{L_s}
\sum_{j=1}^{L_s}
\ell(t_{s,j},X_s).
\label{eq:mse}
\end{equation}
This metric first averages the prediction errors over the future observations of each trajectory and then assigns equal weight to different trajectories. When the future timestamps are regarded as conditional samples drawn from $p(t\mid X)$, the above expression is essentially a Monte Carlo estimator of the following observation-distribution risk:
\begin{equation}
R_{\mathrm{obs}}
=
\mathbb{E}_X
\left[
\int_{\mathcal{T}}
\ell(t,X)p(t\mid X)\,dt
\right]
=
\mathbb{E}_X
\left[
\mathbb{E}_{T\mid X}
[\ell(T,X)]
\right].
\label{eq:obs_risk}
\end{equation}
Therefore, MSE aggregates prediction errors according to the empirical frequency of future timestamps: densely observed temporal regions receive higher weights in the evaluation, whereas sparsely observed regions receive lower weights. In other words, the overall evaluation objective of MSE is determined not only by the model's prediction errors but also by the conditional sampling distribution $p(t\mid X)$.

%% file: paper/method.tex
\section{Continuous-Time Risk for Irregular Forecasting}
In this section, we first define the continuous-time risk that directly corresponds to the continuous-time forecasting objective. We then construct its empirical estimator, CSE, using importance sampling and analyze the asymptotic properties of CSE relative to MSE. Finally, we introduce the global-time risk and decompose the evaluation discrepancy into two components associated with sample-dependent sampling and population-level temporal nonuniformity.

\subsection{Continuous-Time Risk}
As discussed above, the observation-distribution risk corresponding to MSE is affected by the conditional sampling distribution $p(t\mid X)$, thereby coupling model evaluation with sample-specific sampling mechanisms. A natural way to reduce this dependence is to adopt a common target-time measure across all trajectories. For continuous-time forecasting, an ideal evaluation objective should also measure the discrepancy between the predicted trajectory and the ground-truth process over the entire future interval. We therefore adopt the uniform temporal measure and define
\begin{equation}
R_{\mathrm{ct}}
=
\mathbb E_X
\left[
\frac{1}{|\mathcal T|}
\int_{\mathcal T}
\ell(t,X)\,dt
\right].
\label{eq:ct_risk}
\end{equation}
$R_{\mathrm{ct}}$ assigns equal weight to each unit of time within the future interval and is independent of the sampling distribution of the actual observation times. It therefore characterizes the forecasting capability of a model over the entire continuous-time process.

\subsection{Estimating Continuous-Time Risk}
The test set provides ground-truth observations only at a finite number of future timestamps, making it impossible to directly compute the continuous-time integral over the entire interval $\mathcal T$. To estimate $R_{\mathrm{ct}}$ from the available irregular observations, we use the actual conditional sampling distribution $p(t\mid X)$ as the proposal distribution and rewrite the continuous-time risk through importance sampling as
\begin{equation}
\begin{aligned}
R_{\mathrm{ct}}
&=
\mathbb E_X
\left[
\int_{\mathcal T}
\ell(t,X)
\frac{1}{|\mathcal T|p(t\mid X)}
p(t\mid X)\,dt
\right]\\
&=
\mathbb E_X
\left[
\mathbb E_{T\mid X}
\left[
\ell(T,X)
\frac{1}{|\mathcal T|p(T\mid X)}
\right]
\right].
\end{aligned}
\label{eq:ct_importance}
\end{equation}
Here, $1/(|\mathcal T|p(t\mid X))$ corrects the actual conditional sampling distribution toward the uniform temporal distribution. It reduces the contribution of each observation in densely sampled regions while increasing the contribution of each observation in sparsely sampled regions, allowing each temporal location to contribute to the evaluation according to the length of the time interval it represents.

In practical datasets, the conditional sampling density $p(t\mid X_s)$ is typically unknown. Let $\hat p(t_{s,i}\mid X_s)$ denote the conditional temporal density estimated from the future timestamps of the $s$-th trajectory. Since standard importance sampling may exhibit high finite-sample variance due to extreme weights, we adopt self-normalized importance sampling and define the resulting empirical metric under squared error as the Continuous-Time Squared Error (CSE):
\begin{equation}
\mathrm{CSE}
=
\frac{1}{S}
\sum_{s=1}^{S}
\frac{
\sum_{i=1}^{L_s}
\ell(t_{s,i},X_s)/
\hat p(t_{s,i}\mid X_s)
}{
\sum_{i=1}^{L_s}
1/\hat p(t_{s,i}\mid X_s)
}.
\label{eq:cse}
\end{equation}
The constant $1/|\mathcal T|$ in the uniform temporal density cancels between the numerator and denominator. For each trajectory, a continuous-time error estimate is first obtained through self-normalized weighting, after which these estimates are averaged equally across all test trajectories. The same construction naturally yields the Continuous-time Absolute Error (CAE) under absolute error. In the following, we focus on CSE under squared error.

To reduce the influence of an evaluation timestamp on its own density estimate, we employ leave-one-out Gaussian kernel density estimation:
\begin{equation}
\hat p(t_{s,i}\mid X_s)
=
\frac{1}{(L_s-1)h_s}
\sum_{\substack{j=1\\j\neq i}}^{L_s}
K
\left(
\frac{t_{s,i}-t_{s,j}}{h_s}
\right),
\label{eq:kde}
\end{equation}
where $K(\cdot)$ is the Gaussian kernel function and $h_s$ is the bandwidth associated with the $s$-th trajectory. Excluding the current evaluation point prevents it from introducing additional kernel mass at its own location, thereby yielding a more robust estimate of the local temporal density.

To further demonstrate the advantage of CSE over MSE, we theoretically compare their asymptotic estimation errors with respect to the continuous-time risk and introduce the following assumption and theorem.

\begin{assumption}
Let $\mathcal T$ be a compact interval, and assume that different test trajectories are mutually independent. Conditional on $X_s$, the future timestamps are independently and identically distributed according to $p(t\mid X_s)$. Suppose that the loss function is bounded, the conditional sampling density satisfies $p(t\mid X)\geq\rho>0$, and the density estimator $\hat p(t\mid X)$ is strongly consistent over $\mathcal T$.
\label{ass:cse_consistency}
\end{assumption}

\begin{theorem}[Asymptotic Non-Inferiority of CSE]
Let $\underline{L}=\min_{1\leq s\leq S}L_s$. Under Assumption~\ref{ass:cse_consistency}, as both the number of test trajectories $S$ and the minimum number of future observations per trajectory $\underline{L}$ tend to infinity, the following inequality holds almost surely:
\begin{equation}
\lim_{S,\underline{L}\rightarrow\infty}
\left|
\mathrm{CSE}-R_{\mathrm{ct}}
\right|
\leq
\lim_{S,\underline{L}\rightarrow\infty}
\left|
\mathrm{MSE}-R_{\mathrm{ct}}
\right|.
\label{eq:asymptotic_dominance}
\end{equation}
In particular, equality holds when $p(t\mid X)=1/|\mathcal T|$ almost everywhere.
\label{thm:asymptotic_dominance}
\end{theorem}

The complete proof is provided in the Appendix~A. Theorem~\ref{thm:asymptotic_dominance} shows that, under the stated conditions, the asymptotic estimation error of CSE with respect to the continuous-time risk is no greater than that of MSE. When future observation times are uniformly distributed, the two metrics are asymptotically equivalent. When the observation-distribution risk differs from the continuous-time risk, CSE achieves a strictly smaller asymptotic error.

\subsection{Sources of Evaluation Discrepancy}
\label{sec:gap_decomposition}
Because MSE and CSE converge to the observation-distribution risk and the continuous-time risk, respectively, they may yield different model evaluation results. To further analyze this discrepancy, we introduce the global-time risk and decompose the overall discrepancy into two components arising from sample-dependent sampling and population-level temporal nonuniformity.

Marginalizing the conditional sampling distribution over trajectories yields $p_T(t)=\mathbb E_X[p(t\mid X)]$, which describes the distribution of timestamps at the dataset level. Based on this distribution, we define the global-time risk as
\begin{equation}
R_{\mathrm{global}}
=
\mathbb E_X
\left[
\int_{\mathcal T}
\ell(t,X)p_T(t)\,dt
\right].
\label{eq:global_risk}
\end{equation}
$R_{\mathrm{global}}$ removes the influence of sample-specific sampling mechanisms while retaining the population-level nonuniform temporal coverage of the dataset. In this sense, it preserves the global temporal distribution induced by the data collection process, but eliminates the instance-level irregularity that may bias model evaluation. It therefore serves as an intermediate reference between the observed risk $R_{\mathrm{obs}}$ and the continuous-time risk $R_{\mathrm{ct}}$. Based on this risk, the overall evaluation discrepancy can be decomposed as
\begin{equation}
\begin{aligned}
R_{\mathrm{obs}}-R_{\mathrm{ct}}
&=
\underbrace{
R_{\mathrm{obs}}-R_{\mathrm{global}}
}_{G_{\mathrm{samp}}}
+
\underbrace{
R_{\mathrm{global}}-R_{\mathrm{ct}}
}_{G_{\mathrm{time}}},\\
\text{where}\qquad
G_{\mathrm{samp}}
&=
\int_{\mathcal T}
\operatorname{Cov}_X
\left(
\ell(t,X),p(t\mid X)
\right)\,dt,\\
G_{\mathrm{time}}
&=
\operatorname{Cov}_{U\sim\operatorname{Unif}(\mathcal T)}
\left(
\bar\ell(U),|\mathcal T|p_T(U)
\right),
\end{aligned}
\label{eq:risk_gap}
\end{equation}
where $\bar\ell(t)=\mathbb E_X[\ell(t,X)]$. The detailed derivation is provided in the Appendix~B.

$G_{\mathrm{samp}}$ is determined by the covariance, across trajectories, between the model error $\ell(t,X)$ and the sample-specific sampling density $p(t\mid X)$ at a fixed temporal location. If trajectories with larger errors are more likely to be observed at that time, this term is positive; otherwise, it is negative. When the sampling process is independent of the trajectory, i.e., $p(t\mid X)=p_T(t)$, we have $G_{\mathrm{samp}}=0$. Therefore, $G_{\mathrm{samp}}$ characterizes the effect induced by sample-dependent sampling.

$G_{\mathrm{time}}$ is determined by the covariance, across time, between the average model error $\bar{\ell}(t)$ and the population-level relative sampling density $|\mathcal T|p_T(t)$. If temporal regions with larger average errors have higher population-level sampling densities, this term is positive; otherwise, it is negative. When $p_T(t)=1/|\mathcal T|$, we have $G_{\mathrm{time}}=0$. Therefore, $G_{\mathrm{time}}$ characterizes the actual effect of population-level temporal nonuniformity on the evaluation of the current model.

To estimate the above discrepancies in practice, we first estimate the marginal temporal density using $\hat p_T(t)=\frac{1}{S}\sum_{r=1}^{S}\hat p(t\mid X_r)$. We then use $\hat p_T(t)$ as the target distribution to construct a self-normalized importance-sampling estimator of $R_{\mathrm{global}}$, termed the Global-Time Squared Error (GSE). Its detailed form is provided in the Appendix~C. The two discrepancy components can then be estimated as
\begin{equation}
\widehat G_{\mathrm{samp}}
=
\mathrm{MSE}-\mathrm{GSE},
\qquad
\widehat G_{\mathrm{time}}
=
\mathrm{GSE}-\mathrm{CSE}.
\label{eq:empirical_gaps}
\end{equation}
This decomposition enables us to separately quantify the effects of sample-dependent sampling and population-level temporal nonuniformity on model scores and rankings in practice, providing a basis for analyzing evaluation discrepancies in the subsequent experiments.

\section{Benchmark Protocol}
This section presents the benchmark setup, including the synthetic, semi-synthetic, and real-world datasets, representative baseline models, and unified training and evaluation protocols adopted to ensure fair and consistent comparisons.

\subsection{Datasets}
\paragraph{Synthetic and Semi-Synthetic Data.}
To obtain directly computable continuous-time risks, we construct two fully synthetic datasets and two semi-synthetic datasets. Synthetic-Regime gradually transitions from a smooth, low-amplitude regime to a high-frequency periodic regime, simulating a process whose forecasting difficulty varies substantially over time. Synthetic-Multiscale simultaneously contains trends, multiscale periodic patterns, and local transient variations, covering more complex temporal patterns. Both fully synthetic datasets provide analytical continuous ground truth that can be queried at arbitrary future times. The semi-synthetic datasets are constructed from the regularly sampled ETTm1 and Weather datasets~\cite{DBLP:conf/nips/WuXWL21}. Specifically, we sample a finite number of historical and future timestamps from complete regular windows to form irregular samples, and use the equally weighted error over the complete future grid as the reference continuous-time risk. Training target times uniformly cover the future interval, whereas test timestamps are sampled from
\begin{equation}
p_\alpha(u)
=
\alpha\operatorname{Beta}(u;10,60)
+
(1-\alpha)\operatorname{Uniform}(0,1),
\label{eq:test_sampling}
\end{equation}
where $\alpha\in\{0,0.3,0.5,0.9,0.99\}$ controls the degree of temporal nonuniformity. All sampling conditions share the same training data and model predictions, while only the test-time distribution is varied, thereby isolating the effect of the evaluation protocol itself. Detailed data-generation procedures and window configurations are provided in the Appendix~D.1.

\paragraph{Real-World Datasets.}
We select eight real-world irregular time series datasets spanning different application domains and sampling patterns. USHCN contains long-term climate observations from weather stations across the United States~\cite{menne2015ushcn}. MIMIC-III consists of sparse clinical records from ICU patients~\cite{johnson2016mimic}. HumanActivity contains irregularly collected location-sensor measurements recorded during human activities~\cite{li2025hyperimts}. GDELT records multivariate event sequences triggered by international events~\cite{Leetaru13gdelt:global}. RepoHealth describes software repository development activities, whose sampling frequency varies with project activity~\cite{chang2025timeimm}. StudentLife contains mobile sensing data driven by user behavior and daily routines~\cite{nepal2024capturing}. FNSPID consists of stock prices and trading information and exhibits structured temporal gaps induced by trading sessions~\cite{chang2025timeimm}. CESNET contains traffic records generated by network devices, with timestamps affected by system scheduling delays and logging jitter~\cite{chang2025timeimm}. For all datasets, we use only numerical time series and follow the corresponding public settings for preprocessing and data splitting. Detailed dataset statistics, preprocessing procedures, and window configurations are provided in the Appendix~D.2.

\subsection{Baselines}
We select eleven representative irregular time series models covering different temporal modeling paradigms. These include recurrent or continuous-time models, namely GRU-D and NeuralFlow~\cite{che2016grud,bilos2021neuralflows}; set-, attention-, or Transformer-based models, including SeFT, mTAN, and Warpformer~\cite{horn2020seft,shukla2021mtan,zhang2023warpformer}; graph- or hypergraph-based models, including GraFITi, HyperIMTS, tPatchGNN, and ASTGI~\cite{yalavarthi2024grafiti,li2025hyperimts,zhang2024tpatchgnn,liu2026astgi}; and pre-alignment and patch-aggregation methods, including KAFNet and APN~\cite{zhou2026kafnet,liu2026apn}. Detailed descriptions and hyperparameter configurations for all models are provided in the Appendix~D.3.

\subsection{Training and Evaluation Protocol}

All models use the same data splits and input information, and their training objective is uniformly set to the MSE over observed timestamps. The remaining training and hyperparameter settings follow the corresponding public implementations, with model selection performed according to validation MSE. In each independent run, every model produces only one set of test predictions, on which MSE, GSE, and CSE are computed. This ensures that the differences among the metrics arise solely from the temporal weighting applied during test-time evaluation. Both the conditional and marginal temporal densities are estimated exclusively from the test timestamps and are shared across all models on the same dataset. We further compute $\widehat G_{\mathrm{samp}}$ and $\widehat G_{\mathrm{time}}$ to analyze the sources of evaluation discrepancies. All experiments are independently repeated using five random seeds. The main text reports the mean results, while complete results with standard deviations are provided in the Appendix. Detailed training configurations, density-estimation parameters, and other implementation details are also provided in the Appendix~D.4.

%% file: paper/experiments.tex
\section{Experiments}
Using the proposed benchmark, we conduct experiments to investigate three questions:
(1) Can CSE recover continuous-time risk and model rankings more accurately than MSE under non-uniform sampling?
(2) Does observation-point MSE fully reflect models' continuous-time predictive performance on real-world datasets?
(3) How do the sampling-dependence and temporal-distribution gaps influence MSE--CSE discrepancies?

\subsection{Can CSE Better Recover Continuous-Time Risk and Rankings than MSE?}
In this section, we evaluate the ability of CSE to recover continuous-time risks and model rankings in controlled environments with complete future ground truth. The experiments include two fully synthetic datasets and two semi-synthetic datasets, and consider five representative models: SeFT, GRU-D, GraFITi, tPatchGNN, and KAFNet. For the fully synthetic datasets, we compute $R_{\mathrm{ct}}$ through dense temporal integration. For the semi-synthetic datasets, we compute the reference risk over the complete regular future grid. The main text reports the results on Synthetic-Regime and Weather, while the remaining results are provided in the Appendix~E.1.

\begin{table*}[t]
\centering
\small
\setlength{\tabcolsep}{0.9pt}
\renewcommand{\arraystretch}{1.35}
\caption{Forecasting performance under MSE and CSE on eight real-world datasets. Superscripts denote ranks. The best and second-best results are highlighted in bold and underlined, respectively.}
\begin{adjustbox}{max width=\textwidth}
\begin{tabular}{l*{8}{cc}c}
\toprule
& \multicolumn{2}{c}{\shortstack{CESNET\\$\times10^{-1}$}}
& \multicolumn{2}{c}{\shortstack{FNSPID\\$\times10^{-1}$}}
& \multicolumn{2}{c}{\shortstack{GDELT\\$\times10^{0}$}}
& \multicolumn{2}{c}{\shortstack{HumanActivity\\$\times10^{-2}$}}
& \multicolumn{2}{c}{\shortstack{MIMIC-III\\$\times10^{-1}$}}
& \multicolumn{2}{c}{\shortstack{RepoHealth\\$\times10^{-1}$}}
& \multicolumn{2}{c}{\shortstack{StudentLife\\$\times10^{-1}$}}
& \multicolumn{2}{c}{\shortstack{USHCN\\$\times10^{-1}$}}
& \multicolumn{1}{c}{Average} \\
\cmidrule(lr){2-3}
\cmidrule(lr){4-5}
\cmidrule(lr){6-7}
\cmidrule(lr){8-9}
\cmidrule(lr){10-11}
\cmidrule(lr){12-13}
\cmidrule(lr){14-15}
\cmidrule(lr){16-17}
Model
& MSE & CSE
& MSE & CSE
& MSE & CSE
& MSE & CSE
& MSE & CSE
& MSE & CSE
& MSE & CSE
& MSE & CSE
& Rank \\
\midrule

APN
& $8.76\rk{5}$ & $8.58\rk{5}$
& $\underline{2.38}\rk{2}$ & $\underline{2.38}\rk{2}$
& $\mathbf{1.02}\rk{1}$ & $\mathbf{0.99}\rk{1}$
& $5.68\rk{6}$ & $5.65\rk{6}$
& $6.66\rk{6}$ & $6.72\rk{6}$
& $\underline{3.36}\rk{2}$ & $\mathbf{3.57}\rk{1}$
& $\underline{6.61}\rk{2}$ & $6.68\rk{3}$
& $6.04\rk{4}$ & $6.71\rk{5}$
& $\underline{3.6}$ \\

ASTGI
& $\mathbf{8.47}\rk{1}$ & $\mathbf{8.26}\rk{1}$
& $3.28\rk{6}$ & $3.25\rk{6}$
& $1.10\rk{4}$ & $1.04\rk{4}$
& $5.46\rk{3}$ & $5.41\rk{3}$
& $10.50\rk{11}$ & $10.62\rk{11}$
& $6.27\rk{7}$ & $6.53\rk{7}$
& $6.65\rk{3}$ & $\underline{6.65}\rk{2}$
& $5.14\rk{3}$ & $5.66\rk{3}$
& $4.7$ \\

GRU-D
& $10.22\rk{10}$ & $9.91\rk{10}$
& $5.24\rk{8}$ & $5.20\rk{8}$
& $1.22\rk{10}$ & $1.15\rk{10}$
& $8.56\rk{9}$ & $8.56\rk{9}$
& $6.67\rk{7}$ & $6.75\rk{7}$
& $11.68\rk{11}$ & $11.95\rk{11}$
& $7.21\rk{8}$ & $7.23\rk{8}$
& $7.78\rk{11}$ & $8.59\rk{11}$
& $9.3$ \\

GraFITi
& $\underline{8.50}\rk{2}$ & $\underline{8.28}\rk{2}$
& $3.21\rk{4}$ & $3.22\rk{4}$
& $1.09\rk{3}$ & $\underline{1.01}\rk{2}$
& $\underline{5.44}\rk{2}$ & $\underline{5.40}\rk{2}$
& $\mathbf{5.89}\rk{1}$ & $\underline{6.02}\rk{2}$
& $\mathbf{3.32}\rk{1}$ & $\underline{3.61}\rk{2}$
& $6.72\rk{6}$ & $6.70\rk{4}$
& $\mathbf{4.50}\rk{1}$ & $\mathbf{4.89}\rk{1}$
& $\mathbf{2.4}$ \\

HyperIMTS
& $8.90\rk{7}$ & $8.65\rk{7}$
& $2.72\rk{3}$ & $2.71\rk{3}$
& $1.18\rk{8}$ & $1.09\rk{8}$
& $\mathbf{5.42}\rk{1}$ & $\mathbf{5.38}\rk{1}$
& $\underline{5.94}\rk{2}$ & $\mathbf{5.92}\rk{1}$
& $3.72\rk{3}$ & $3.97\rk{3}$
& $\mathbf{6.54}\rk{1}$ & $\mathbf{6.56}\rk{1}$
& $6.67\rk{8}$ & $7.30\rk{8}$
& $4.1$ \\

KAFNet
& $8.73\rk{4}$ & $8.54\rk{4}$
& $\mathbf{2.23}\rk{1}$ & $\mathbf{2.22}\rk{1}$
& $\underline{1.08}\rk{2}$ & $1.02\rk{3}$
& $5.62\rk{4}$ & $5.58\rk{4}$
& $6.41\rk{3}$ & $6.47\rk{3}$
& $5.72\rk{6}$ & $5.98\rk{6}$
& $6.66\rk{4}$ & $6.72\rk{5}$
& $7.14\rk{9}$ & $7.82\rk{9}$
& $4.3$ \\

NeuralFlow
& $9.98\rk{8}$ & $9.69\rk{8}$
& $7.33\rk{10}$ & $7.29\rk{10}$
& $1.18\rk{9}$ & $1.09\rk{9}$
& $17.53\rk{11}$ & $17.51\rk{11}$
& $7.05\rk{8}$ & $7.19\rk{9}$
& $10.13\rk{10}$ & $10.44\rk{10}$
& $7.83\rk{10}$ & $7.86\rk{10}$
& $6.07\rk{5}$ & $6.74\rk{6}$
& $9.0$ \\

SeFT
& $10.11\rk{9}$ & $9.79\rk{9}$
& $8.70\rk{11}$ & $8.66\rk{11}$
& $1.13\rk{6}$ & $1.05\rk{5}$
& $9.38\rk{10}$ & $9.36\rk{10}$
& $7.10\rk{9}$ & $7.15\rk{8}$
& $9.13\rk{8}$ & $9.44\rk{8}$
& $8.94\rk{11}$ & $8.95\rk{11}$
& $7.64\rk{10}$ & $8.41\rk{10}$
& $9.1$ \\

Warpformer
& $8.58\rk{3}$ & $8.37\rk{3}$
& $3.66\rk{7}$ & $3.63\rk{7}$
& $1.30\rk{11}$ & $1.20\rk{11}$
& $5.65\rk{5}$ & $5.61\rk{5}$
& $6.42\rk{4}$ & $6.50\rk{5}$
& $5.04\rk{5}$ & $5.33\rk{5}$
& $6.78\rk{7}$ & $6.80\rk{7}$
& $6.13\rk{6}$ & $6.67\rk{4}$
& $5.9$ \\

mTAN
& $10.47\rk{11}$ & $10.14\rk{11}$
& $6.13\rk{9}$ & $6.09\rk{9}$
& $1.12\rk{5}$ & $1.05\rk{6}$
& $7.33\rk{8}$ & $7.30\rk{8}$
& $7.47\rk{10}$ & $7.54\rk{10}$
& $10.09\rk{9}$ & $10.39\rk{9}$
& $7.32\rk{9}$ & $7.34\rk{9}$
& $\underline{4.84}\rk{2}$ & $\underline{5.23}\rk{2}$
& $7.9$ \\

tPatchGNN
& $8.86\rk{6}$ & $8.61\rk{6}$
& $3.26\rk{5}$ & $3.24\rk{5}$
& $1.15\rk{7}$ & $1.08\rk{7}$
& $5.80\rk{7}$ & $5.76\rk{7}$
& $6.44\rk{5}$ & $6.48\rk{4}$
& $4.55\rk{4}$ & $4.82\rk{4}$
& $6.68\rk{5}$ & $6.73\rk{6}$
& $6.36\rk{7}$ & $7.05\rk{7}$
& $5.8$ \\

\bottomrule
\end{tabular}
\end{adjustbox}
\label{tab:mse_cse_all_models}
\end{table*}

\begin{figure}
    \centering
    \begin{subfigure}[b]{0.23\textwidth} 
        \centering
        \includegraphics[width=\textwidth]{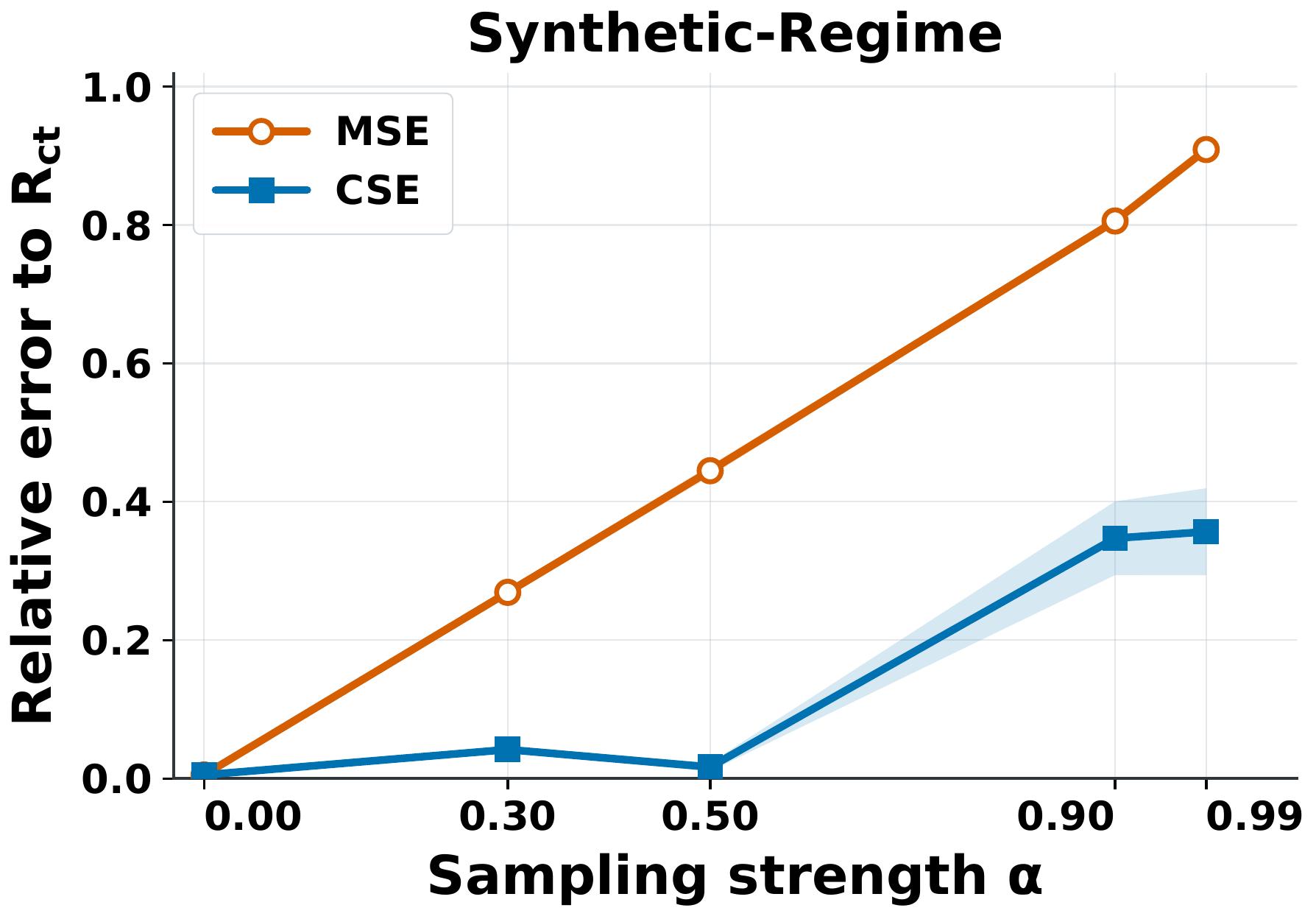} 
        \caption{Synthetic-Regime}
    \end{subfigure}
    \hfill 
    \begin{subfigure}[b]{0.23\textwidth}
        \centering
        \includegraphics[width=\textwidth]{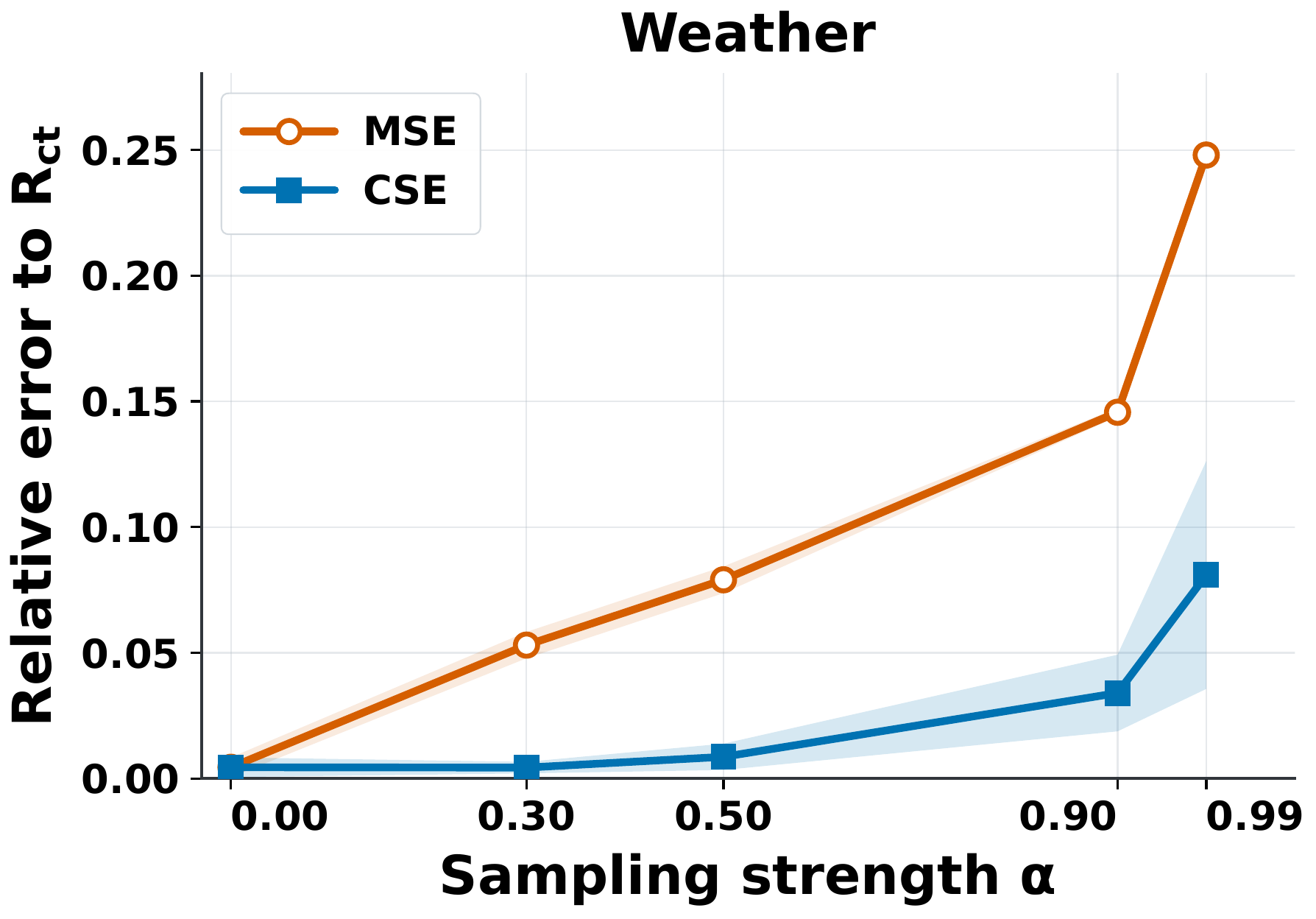}
        \caption{Weather}
    \end{subfigure}
    \caption{Relative errors of MSE and CSE with respect to $R_{\mathrm{ct}}$ under increasing sampling non-uniformity.}
    \label{fig:recovery}
\end{figure}

Fig.~\ref{fig:recovery}(a) and Fig.~\ref{fig:recovery}(b) show the estimation errors of MSE and CSE with respect to $R_{\mathrm{ct}}$. Under uniform sampling, both metrics accurately recover the continuous-time risk. As $\alpha$ increases, the test-time distribution progressively deviates from the uniform distribution, causing the discrepancy between MSE and $R_{\mathrm{ct}}$ to increase consistently, whereas CSE maintains a substantially lower estimation error. For example, when $\alpha=0.9$, the relative errors of MSE and CSE on Synthetic-Regime are $80.6\%$ and $34.7\%$, respectively. Under the more extreme setting of $\alpha=0.99$, although the error of CSE increases, it remains substantially lower than that of MSE.

To further assess the recovery of model rankings, Table~\ref{tab:ct_risk_recovery} compares the evaluation results of the five models under $\alpha=0.9$. On both Synthetic-Regime and Weather, CSE recovers exactly the same ranking of all five models as $R_{\mathrm{ct}}$, whereas MSE produces a different ranking on both datasets.

Overall, CSE effectively mitigates the influence of non-uniform test sampling on model evaluation. Across both synthetic and semi-synthetic datasets, CSE recovers continuous-time risk and model rankings more accurately than MSE, validating its effectiveness as a continuous-time evaluation metric.

\begin{table}[t]
\centering
\setlength{\tabcolsep}{1.5pt}
\renewcommand{\arraystretch}{1.25}
\caption{Model evaluation under non-uniform test sampling with $\alpha=0.9$. Superscripts denote ranks; the best and second-best results are highlighted in bold and underlined, respectively.}
\begin{adjustbox}{max width=\linewidth}
\begin{tabular}{lccc ccc}
\toprule
& \multicolumn{3}{c}{\shortstack{Synthetic-Regime\\$\times 10^{-2}$}}
& \multicolumn{3}{c}{\shortstack{Weather\\$\times 10^{-1}$}} \\
\cmidrule(lr){2-4}
\cmidrule(lr){5-7}
Model
& MSE
& CSE
& $R_{\mathrm{ct}}$
& MSE
& CSE
& $R_{\mathrm{ct}}$ \\
\midrule

SeFT
& $1.09\rk{3}$
& $\mathbf{8.70}\rk{1}$
& $\mathbf{6.26}\rk{1}$
& $4.40\rk{5}$
& $4.96\rk{5}$
& $4.85\rk{5}$ \\

GRU-D
& $\mathbf{0.65}\rk{1}$
& $\underline{8.81}\rk{2}$
& $\underline{6.48}\rk{2}$
& $\mathbf{3.08}\rk{1}$
& $4.61\rk{3}$
& $4.46\rk{3}$ \\

GraFITi
& $2.57\rk{5}$
& $9.88\rk{5}$
& $7.66\rk{5}$
& $3.89\rk{4}$
& $4.92\rk{4}$
& $4.76\rk{4}$ \\

tPatchGNN
& $\underline{0.91}\rk{2}$
& $9.38\rk{4}$
& $6.91\rk{4}$
& $3.88\rk{3}$
& $\mathbf{3.69}\rk{1}$
& $\mathbf{3.53}\rk{1}$ \\

KAFNet
& $1.54\rk{4}$
& $9.04\rk{3}$
& $6.75\rk{3}$
& $\underline{3.48}\rk{2}$
& $\underline{3.78}\rk{2}$
& $\underline{3.65}\rk{2}$ \\

\bottomrule
\end{tabular}
\end{adjustbox}
\label{tab:ct_risk_recovery}
\end{table}

\begin{figure*}
    \centering
    \begin{subfigure}[b]{0.49\textwidth} 
        \centering
        \includegraphics[width=\textwidth]{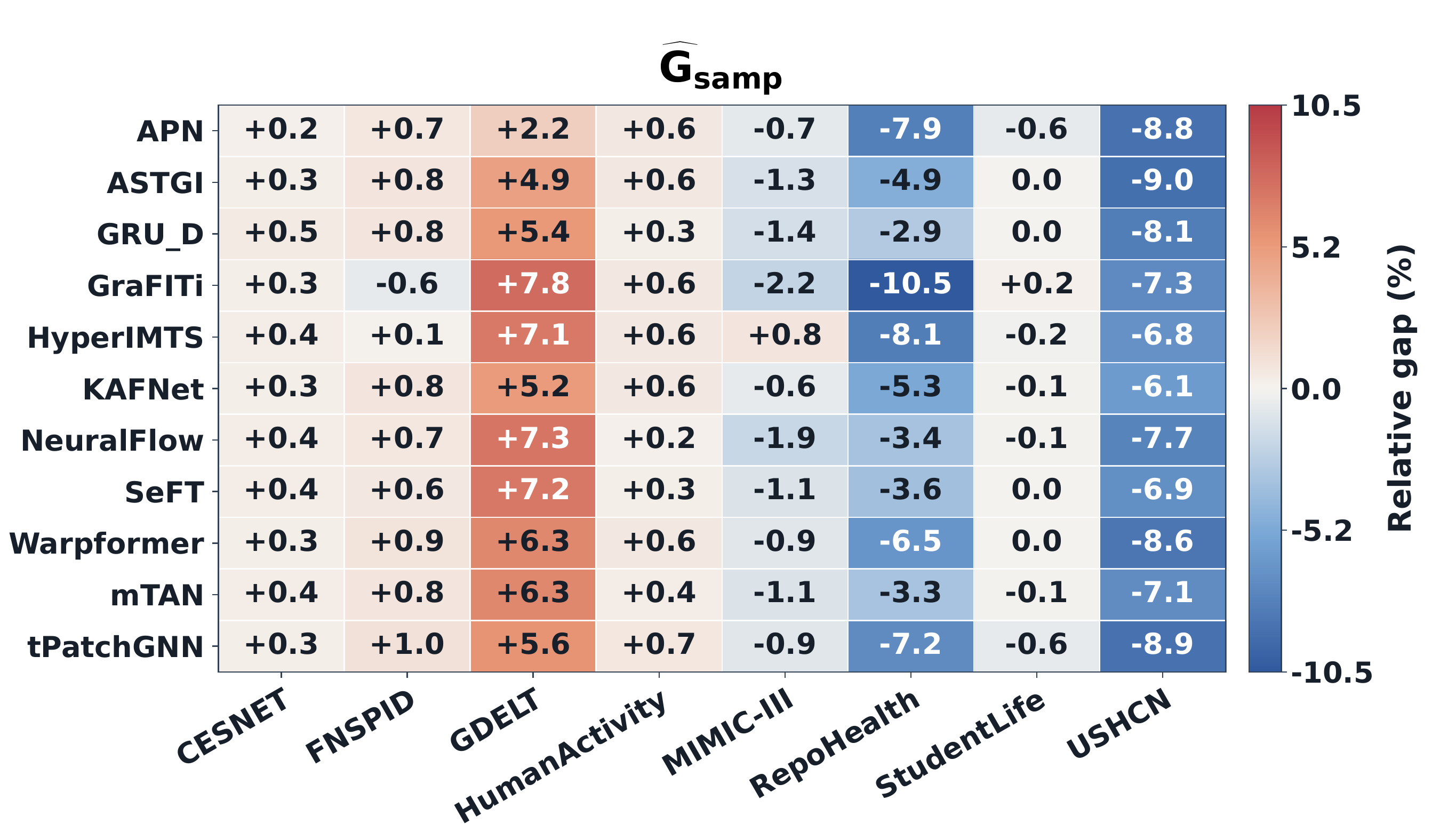} 
        \caption{Sampling-dependence gap $\widehat G_{\mathrm{samp}}$}
    \end{subfigure}
    \hfill 
    \begin{subfigure}[b]{0.49\textwidth}
        \centering
        \includegraphics[width=\textwidth]{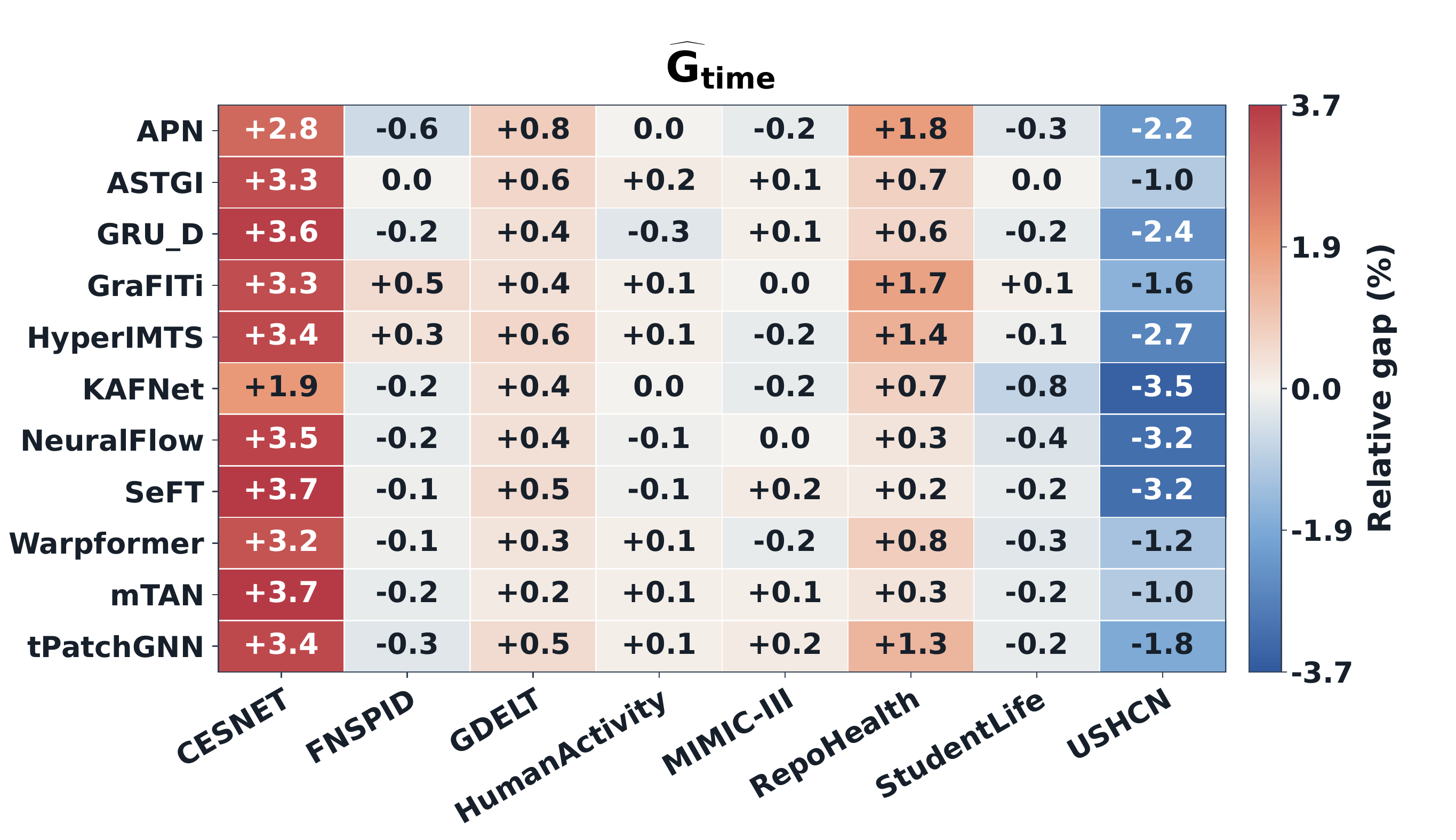}
        \caption{Temporal-distribution gap $\widehat G_{\mathrm{time}}$}
    \end{subfigure}
    \caption{Normalized decomposition of the evaluation discrepancy into the sampling-dependence gap $\widehat G_{\mathrm{samp}}$ and temporal-distribution gap $\widehat G_{\mathrm{time}}$ across models and datasets.}
    \label{fig:exp3}
\end{figure*}

\subsection{Does MSE Fully Reflect Continuous-Time Predictive Performance on Real-World Data?}

\begin{figure}
    \centering
    \begin{subfigure}[b]{0.23\textwidth} 
        \centering
        \includegraphics[width=\textwidth]{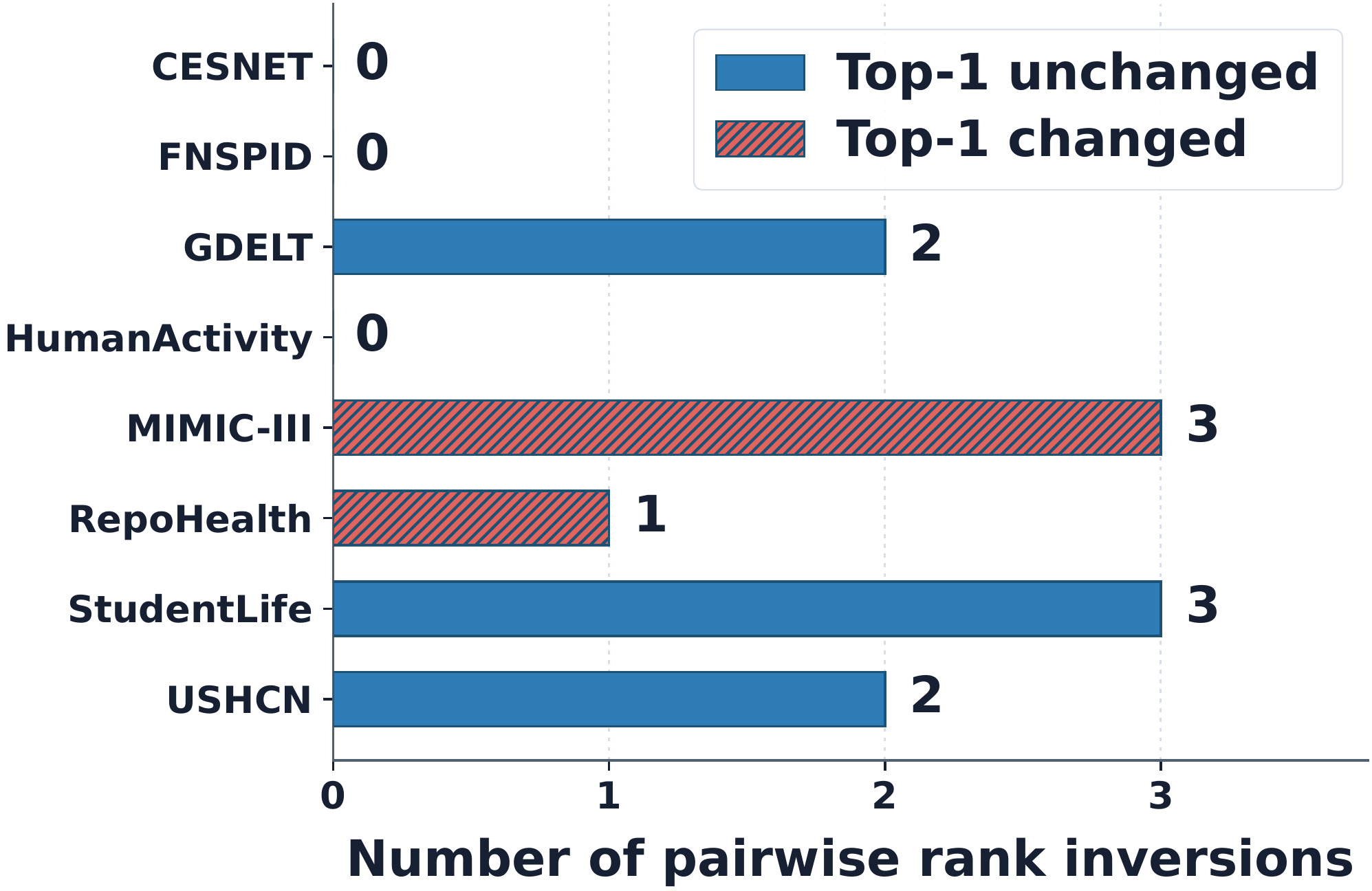} 
        \caption{Pairwise rank inversions across datasets}
    \end{subfigure}
    \hfill 
    \begin{subfigure}[b]{0.23\textwidth}
        \centering
        \includegraphics[width=\textwidth]{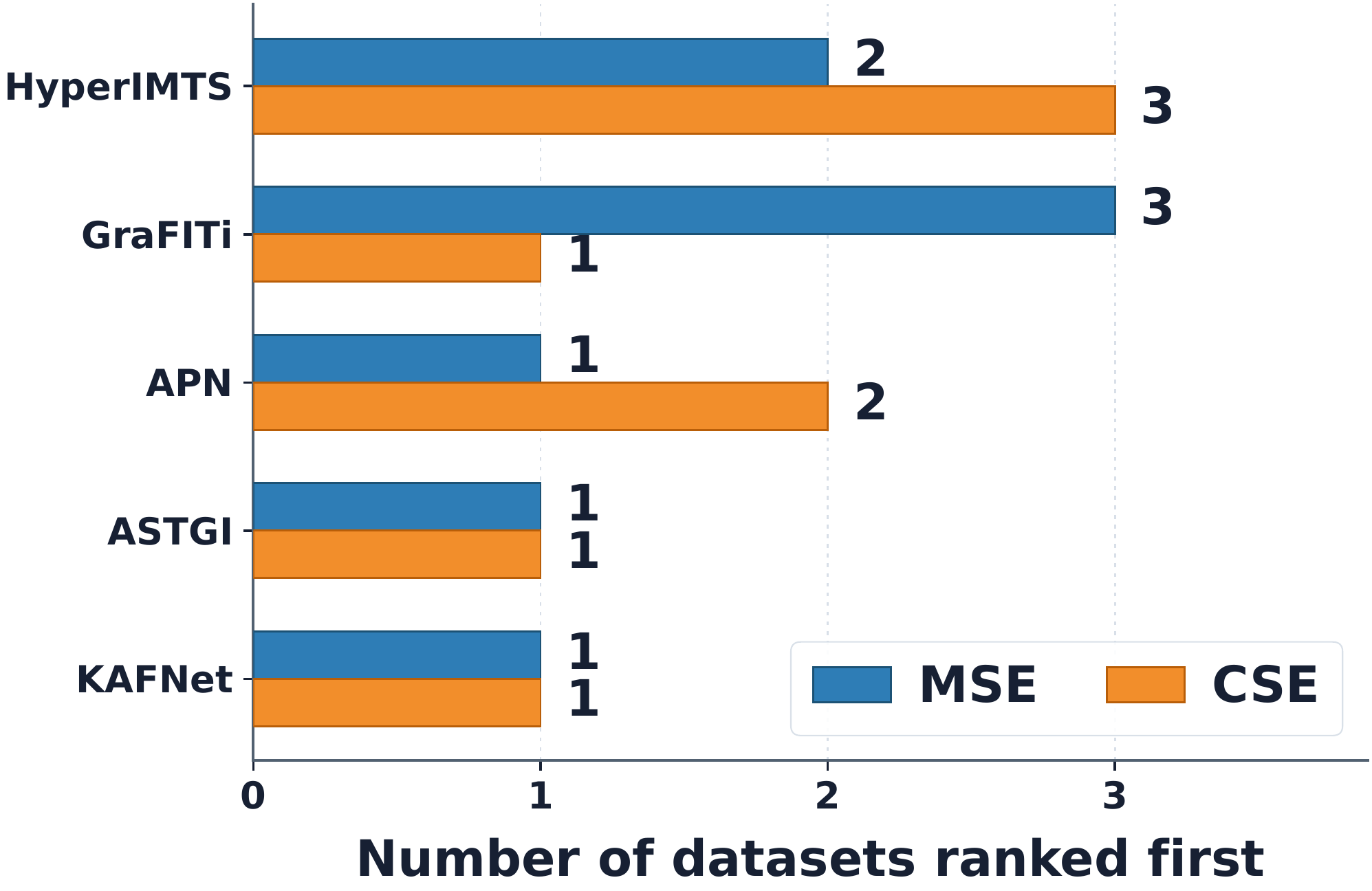}
        \caption{Changes in Top-1 model selection}
    \end{subfigure}
    \caption{Model ranking changes when replacing MSE with CSE on eight real-world datasets. Panel (a) reports the number of pairwise rank inversions, while Panel (b) compares the frequency with which each model is ranked first.}
    \label{fig:exp2}
\end{figure}

The previous section shows that CSE more accurately recovers continuous-time risk and model rankings when the underlying future trajectory is available. We now examine whether observation-point MSE provides a sufficient evaluation of models' continuous-time predictive performance on real-world datasets, where the complete continuous trajectory is unavailable. Table~\ref{tab:mse_cse_all_models} reports the results of 11 models on eight datasets, while Fig.~\ref{fig:exp2} summarizes pairwise rank inversions and changes in the selection of the best-performing model. Comparisons between MAE and CAE, together with complete results including standard deviations, are provided in the Appendix~E.2.

The comparison shows that CSE is not a fixed-direction or fixed-ratio correction of MSE. The direction and magnitude of score changes vary across both datasets and models, indicating that the two metrics cannot be converted through simple rescaling. Moreover, five of the eight datasets exhibit pairwise rank inversions when replacing MSE with CSE. MIMIC-III and StudentLife each contain three inversions, GDELT and USHCN each contain two, and RepoHealth contains one. The best-performing model also changes on MIMIC-III and RepoHealth.

The effects further differ across models. GraFITi ranks first on three datasets under MSE but on only one dataset under CSE. In contrast, the number of first-place rankings increases from two to three for HyperIMTS and from one to two for APN. Although GraFITi retains the best average rank, MSE and CSE do not provide fully consistent conclusions regarding the relative advantages of different models.

Overall, the substantial discrepancies in model scores, rankings, and model selection suggest that relying solely on observation-point MSE may not fully reflect models' continuous-time predictive performance on real-world datasets.

\subsection{How Do the Sampling-Dependence and Temporal-Distribution Gaps Influence MSE--CSE Discrepancies?}

The previous section demonstrates that MSE and CSE can produce different model scores, rankings, and model selections on real-world datasets. We now use the risk decomposition introduced above to analyze the sources of these discrepancies. Fig.~\ref{fig:exp3} reports the normalized sampling-dependence gap,
$\widehat{G}_{\mathrm{samp}}=(\mathrm{MSE}-\mathrm{GSE})/\mathrm{MSE}$,
and the normalized temporal-distribution gap,
$\widehat{G}_{\mathrm{time}}=(\mathrm{GSE}-\mathrm{CSE})/\mathrm{GSE}$.

The relative contributions of the two gaps vary substantially across datasets. GDELT and RepoHealth are primarily affected by the sampling-dependence gap, whose cross-model averages are approximately $5.9\%$ and $-5.8\%$, respectively. CESNET is mainly affected by the temporal-distribution gap, with an average value of approximately $3.3\%$. On USHCN, the sampling-dependence and temporal-distribution gaps are approximately $-7.8\%$ and $-2.2\%$, respectively, indicating that both components contribute to the discrepancy between MSE and CSE. The two gaps are comparatively small on the remaining datasets.

Within the same dataset, models generally exhibit similar signs for each discrepancy component, suggesting that the overall direction is largely determined by the dataset's sampling structure. However, the magnitude varies across models because of their different temporal error patterns. This model-dependent effect can alter model rankings when the original performance differences are small. For example, on MIMIC-III, the sampling-dependence gaps of GraFITi and HyperIMTS are $-2.2\%$ and $+0.8\%$, respectively, changing the best-performing model from GraFITi under MSE to HyperIMTS under CSE. On RepoHealth, GraFITi exhibits a more negative sampling-dependence gap than APN, giving it a larger apparent advantage under MSE; this advantage disappears under CSE, making APN the best-performing model.

Overall, the discrepancies between MSE and CSE are jointly influenced by the sampling-dependence and temporal-distribution gaps, whose relative contributions vary across datasets and models.

%% file: paper/conclusion.tex
\section{Conclusion}
This work revisits the evaluation of irregular time series forecasting from a continuous-time perspective. We show that observation-point MSE and MAE effectively estimate the observation-distribution risk induced by the sampling distribution, rather than the continuous-time risk over the entire future interval. To mitigate this objective mismatch, we propose the Continuous-Time Squared Error (CSE), which estimates the continuous-time risk through self-normalized importance weighting, and prove that its asymptotic estimation error is no greater than that of MSE. We further decompose the discrepancy between the two risks into a sampling-dependence gap and a temporal-distribution gap, thereby characterizing how different sampling structures affect model evaluation. 
Finally, we construct a benchmark covering synthetic, semi-synthetic, and real-world datasets to validate CSE and systematically evaluate models' continuous-time predictive performance. 
